\documentclass[11pt]{article}

\usepackage[final]{acl}

\usepackage{times}
\usepackage{latexsym}
\usepackage{placeins}
\usepackage[T1]{fontenc}
\usepackage{seqsplit}
\usepackage{fontawesome5}
\usepackage{hyperref}
\usepackage[utf8]{inputenc}
\usepackage{booktabs}
\usepackage{multirow}
\usepackage{float}
\usepackage[normalem]{ulem}
\useunder{\uline}{\ul}{}
\usepackage{microtype}
\usepackage{amsmath, amssymb}
\usepackage{mathtools}
\usepackage{booktabs}
\usepackage{geometry}
\usepackage{titlesec}
\usepackage{microtype}
\usepackage{longtable}
\usepackage{colortbl}
\usepackage{xcolor}
\usepackage{array}
\usepackage{subcaption} %

\usepackage{amsmath,amsfonts,bm,mathtools,amssymb}
\usepackage{makecell}
\usepackage{stfloats}
\usepackage{inconsolata}

\usepackage{graphicx}

\title{Reading Time Modeling with Pixel-Based Language Models}
\author{Junlin Li$^{1}$
  \quad
  Sara Alhajeri$^{1}$
  \quad
  Tatsuki Kuribayashi$^{1,2}$ \\
$^1$Mohamed bin Zayed University of Artificial Intelligence \quad
$^2$Tohoku University \\
    \texttt{\{junlin.li, sara.alhajeri, tatsuki.kuribayashi\}@mbzuai.ac.ae}\\
    }

\begin{document}
\maketitle
\begin{abstract}
Tokenization-free, pixel-based language models (PLMs) have recently been proposed as an alternative to conventional token-based language models (TLMs), typically motivated by practical considerations, e.g., the flexibility to support new languages or vocabularies.
Nevertheless, PLMs also have the potential to serve as infrastructure for computational (psycho)linguistics, where tokenization has been a source of bias in estimating information-theoretic values, such as surprisal, and has pushed the bottom-up visual orthographic features outside the scope.
In this study, we first integrate PLMs into the modeling of naturalistic human reading and, in particular, examine their use to estimate next-word predictability, which can also integrate orthographic and parafoveal information.

Our results demonstrate that next-word prediction difficulty, computed in \textbf{pixel space}, has strong predictive power for gaze duration and word skipping behavior across various languages. In particular, their advantage is complementary to existing token-level surprisal and entropy estimates.
Cross-lingual analyses with script variations also clarify their distinctive visual processing biases, facilitating research into how bottom-up visual information is incrementally integrated into expectation-based reading. 
\newline
\newline
\noindent
\faGithub\hspace{0.3em}
\href{https://github.com/CN-Eyetk/reading-time-modeling-with-pixel-lm}
{\texttt{reading-time-modeling-with-pixel-lm}}
\end{abstract}

\section{Introduction}

Reading coordinates a stratification of low- and high-level cognitive processes \cite{engbert2005swift} across multiple modalities \cite{hoover1990simple}. 
Naturalistic reading data, for example, indicate that a considerable proportion of words is processed without a fixation; they are possibly recognized in the parafovea \cite{schotter2012parafoveal, rayner1998eye} with only partial visual information \cite{reichle2003ez}. 
Such a rapid pre-attentive visual processing suggests the importance of a bottom-up sensory prediction \cite{kok2015predictive,tarkiainen1999dynamics}, especially how graded visual presentation affects reading time and skipping \cite{gordon2013see} via top-down prediction \cite{ding2025modeling}.

\begin{figure}[t]
    \centering
    \includegraphics[width=\linewidth]{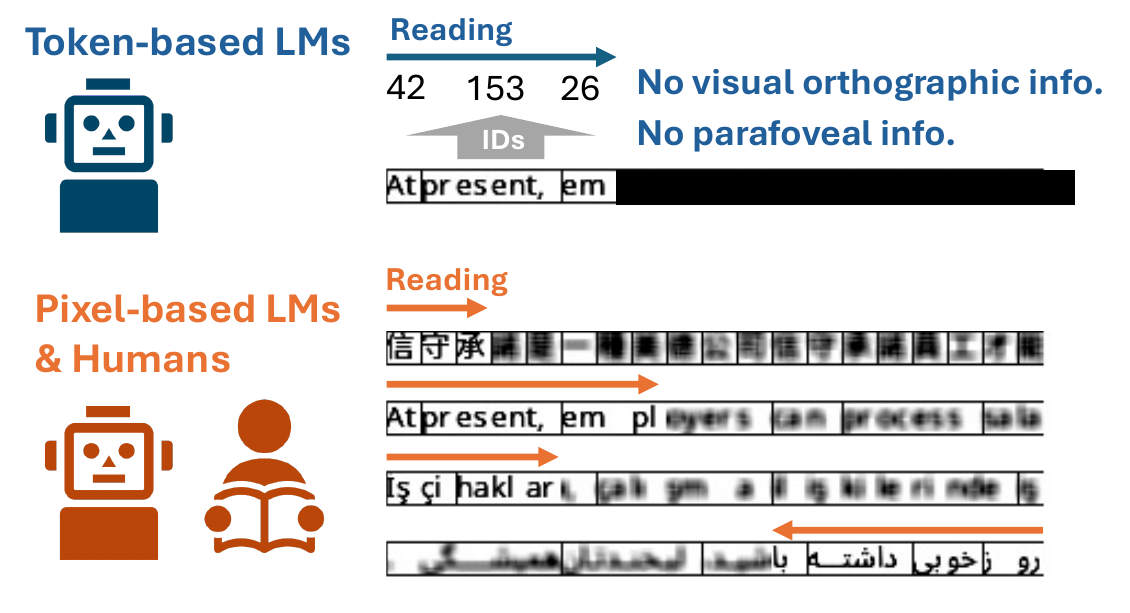}
    \caption{Simulating visual sentence processing with orthographic and parafoveal information using pixel-based LMs}
    \label{fig:placeholder}
\end{figure}

In computational psycholinguistics, neural language models (LMs) have been ubiquitous in estimating information-theoretic values, such as surprisal \cite{hale2001, levy2008expectation}, to instantiate linguistic theories.
Despite their successful interdisciplinary bridge, recent studies have also identified drawbacks of ordinal LMs defined over a fixed set of token IDs (token-based language models; TLMs) as tools for cognitive modeling. 
In particular, their tokenizers and vocabularies, predefined regardless of their cognitive plausibility \cite{haslett2025tokenization}, potentially limit feasible experimental designs, e.g., estimating highly incremental probabilities across more fine-grained units, or implicitly bias experimental results~\cite{nair-resnik-2023-words,oh2024leading,giulianelli-etal-2024-proper,oh-schuler-2025-impact,kiegeland2026proper}. 
For example, \citet{nair-resnik-2023-words} reported that a tokenizer or (sub)word definition easily biases the surprisal estimate $-\log p(\mathrm{word}|\mathrm{context})$.
Another example is that a naive accumulation of subword probabilities results in their misallocation to the following word, e.g., ``He give \underline{me}'', where the load of ``give'' can be dislocated to ``\underline{me},'' if the alternative ``gives'' is also split into the common prefix ``give'' and ``s''~\cite{oh2024leading,giulianelli-etal-2024-proper}.
Although some problems have quick solutions, tokenizer selection is inherently the source of biases in computational psycholinguistic experiments with TLMs.
More fundamentally, TLM also limits the possibility of experimental design; for example, the word-to-ID conversion makes it impossible to study the visual and orthographic effects in reading.

In this paper, to handle such \textit{tokenization bottlenecks} spread in modern TLM-based cognitive modeling, we examine the use of tokenization-free, pixel-based language models (\textbf{PLM}s) \cite{borenstein2023phd, rust2023language, lotz2023text} in computational psycholinguistics.
PLMs encode text as pixel-based images, potentially without loss of visual fidelity to what humans can indeed see, enabling flexible research designs, for example, to explore how text layout, font, or orthographic complexity drives reading \cite{borenstein2023phd}.
This work, as a case study, specifically explores the psychometric predictive power of PLMs and parafoveal effects in reading, positing that early-stage visual processing may already begin before fixating on the exact word, and from an expectation-based reading perspective, visual word identifiability will be a strong predictor of human gaze duration and scan path.
We compute word predictability and entropy with PLMs that rely on orthographic information and a blurred future context, intended to reflect parafoveal sight.
These estimates are used to model reading time and word skipping behavior during naturalistic reading.

Our experiments address the following questions:
\begin{itemize}
    \item \textbf{Q1: } Do PLMs without a tokenizer predict reading time and word skipping better than or complementary to TLM surprisal, which relies on a tokenizer?
    \item \textbf{Q2 }: Does PLM-estimated visual word-prediction difficulty offer a general predictor for cognitive modeling across different languages and writing systems?
\end{itemize}

Our experimental results show that the visual word prediction difficulty estimated by PLMs offers complementary predictive power to conventional token-based surprisal.
This advantage is observed across languages, particularly in those with scripts other than the Latin alphabet, demonstrating that PLM is a useful infrastructure for estimating word predictability and modeling human reading across diverse languages.

\section{Related Work}

\subsection{Reading Time Modeling with Token-based Language Models}
Language models (LMs) have been widely used as tools to implement and quantitatively test (neuro and psycho)linguistic theories~\cite{caucheteux2023evidence,OHAMS2026104705}.
Given that LMs are basically next-word prediction machines, their use is typically relevant to testing an expectation-based account of human sentence processing through estimating information-theoretic measures such as surprisal~\cite{hale2001, levy2008expectation,hale2016information}. Curiously, a misalignment between human reading measures and LM-estimated predictive measures has been indicated by recent studies \cite{kuribayashi-etal-2021-lower,shain2022large,oh2023does}, raising several possible explanations.
A conceptually intuitive implication is that prediction is not a complete representation of human sentence processing \cite{slaats2025s}, which features structural modularistic underpinnings \cite{coltheart2001drc} and thus is not purely probabilistic. But practically, another possibility rests on the biases of the proxies, i.e., LMs, in estimating predictive measures \cite{nguyen2026word}. 
Existing studies~\cite{nair-resnik-2023-words,giulianelli-etal-2024-proper,oh2024leading,oh-schuler-2025-impact}, for example, demonstrate that the tokenizer substantially affects LM-based cognitive modeling. Besides, language prediction based on token identity prevents any bottom-up visual and orthographic decoding processes. This study explores a more neutral, realistic design for text encoding in LM-based cognitive modeling, i.e., simply inputting the text image as we read texts with our eyes.

\subsection{Pixel-Based Language Models}
Fortunately, there are practical techniques for processing language using text images as input.
Pixel-based language model (\textsc{PLM}) \cite{rust2023language}, using a visual Transformer (ViT) as its backbone, is a representative model that seamlessly models language in a tokenizer-free paradigm, regardless of vocabulary and character-inventory constraints~\cite{lotz2023text}. 
Such \textsc{PLM}s have been adopted, for example, to enable a flexible addition of a new language or to process low-resource languages, such as historical languages without transcriptions~\cite{borenstein2023phd}. 
Beneath their powerful performance in the visual domain, probing analyses indicate that \textsc{PLMs} also learn linguistic knowledge, such as semantics and syntax~\cite{tatariya2024pixology} and exhibit cross-script generalizability through multilingual pre-training \cite{kesen2025multilingual}.
In contrast to these studies, we examine the use of PLMs for cognitive modeling.

\begin{figure}
    \centering
    \includegraphics[width=\linewidth]{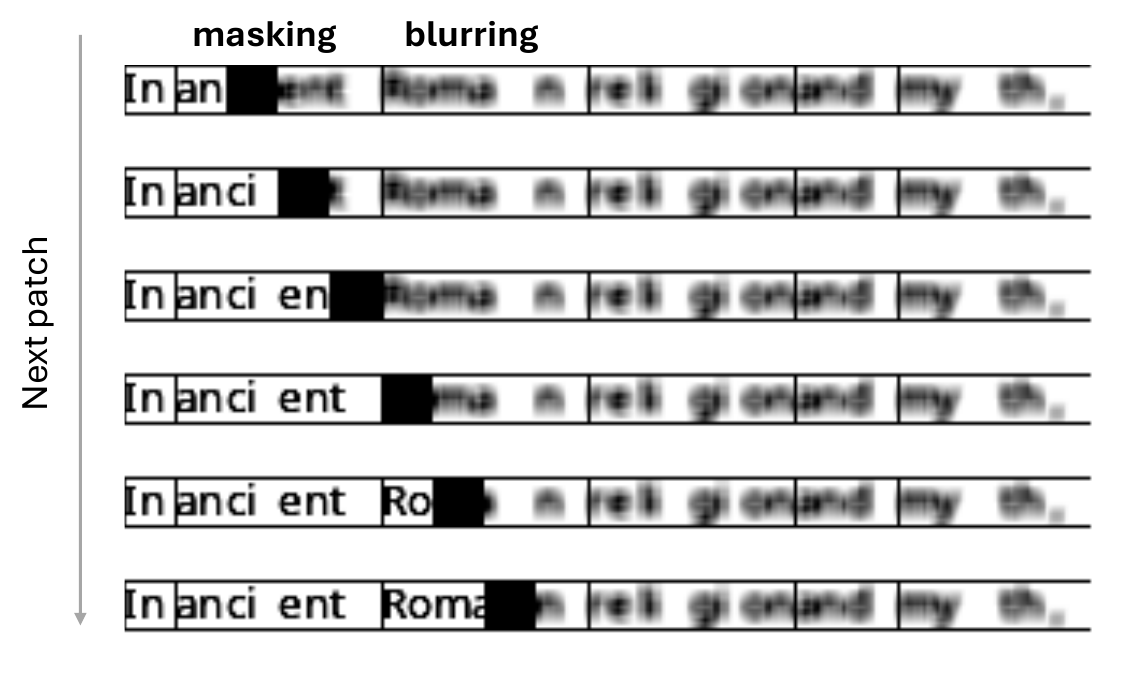}
    \caption{We approximate next-word predictability by patch-by-patch masking and blurring of the future context, simulating parafoveal sight, using \textsc{PLM}.}
    \label{fig:masking}
\end{figure}

\section{Methods}
This section briefly introduces PLMs, which model text sequences in the visual modality without relying on a tokenizer or a discrete vocabulary.
We then explain how to compute next-word predictability with PLMs.

\subsection{Pixel-Based Language Models}
Given a token sequence \(\bm w=[w_1,w_2,\dots,w_n]\), a rendering function converts the sequence into an image
\(\bm x\in\mathbb{R}^{H\times W\times C}\),
where \(C\), \(H\), and \(W\) denote the number of channels (e.g., RGB), height, and width, respectively.
The rendered image is divided into $N = \frac{HW}{P^2}$
non-overlapping patches $\bm X = [\bm x_1, \cdots, \bm x_N] \in \mathbb{R}^{N\times P^2\times C}$, where $P$ is the height/width of each patch. 
Each patch $\bm x_i \in \mathbb{R}^{P^2\times C}$ is flattened and projected into a \(d\)-dimensional embedding $\bm h_i \in \mathbb{R}^d$ with an embedding layer of a PLM:

\begin{equation}
    \bm h_i
=
\mathrm{Flatten}(\bm x_i)\bm W_E
+
\bm b_E
+
\bm e_i^{\mathrm{pos}},
\end{equation}

\noindent
where $\bm W_E \in \mathbb{R}^{P^2C\times d} $ and $\bm b_E \in \mathbb{R}^d$ are learnable parameters, and \(\bm e_i^{\mathrm{pos}} \in \mathbb{R}^d\) denotes the positional embedding of the \(i\)-th patch.
$\bm H=[\bm h_1,\dots,\bm h_N]$ denotes the sequence of patch embeddings.

During training, a subset of patches in $\bm H$ is randomly masked. 
Let indices for masked patches be $\mathcal{I} \subset [1,\cdots,N]$.
The masked input $\bm H_\mathrm{masked}$ is processed by a PLM $\mathcal{F}$ to reconstruct the original, unmasked version of the patches:\footnote{Strictly speaking, the PLM architecture used~\cite{rust2023language,kesen2025multilingual} adopts an encoder-decoder style, where the masked patches are excluded from the encoder input, and learnable mask embeddings are inserted in the decoder part. Note that our claim does not assume specific PLM architectures, and we begin with a unidirectional formulation in Section~\ref{method:1}.}

\begin{align}
\hat{\bm{X}}= \mathcal{F}(\bm{H}_\mathrm{masked})\;\;\mathrm{,}
\end{align}
\noindent
where $\hat{\bm{X}}=\{\hat{\bm{x}}_1,\dots,\hat{\bm{x}}_N\} \in\mathbb{R}^{N\times P^2\times C}$.
Then the reconstruction loss $\mathcal{L}$ is calculated over the indices $\mathcal{I}$ where the input was masked, and basically, the following objective is minimized:

\begin{align}
\mathcal{L}
=
\frac{1}{|\mathcal{I}|}
\sum_{i\in\mathcal{I}}
\left\|
\hat{\bm{x}}_i-\bm{x}_i
\right\|_2^2 \;\;\mathrm{.}
\end{align}

\begin{figure}[t]
    \centering
    \includegraphics[width=0.95\linewidth]{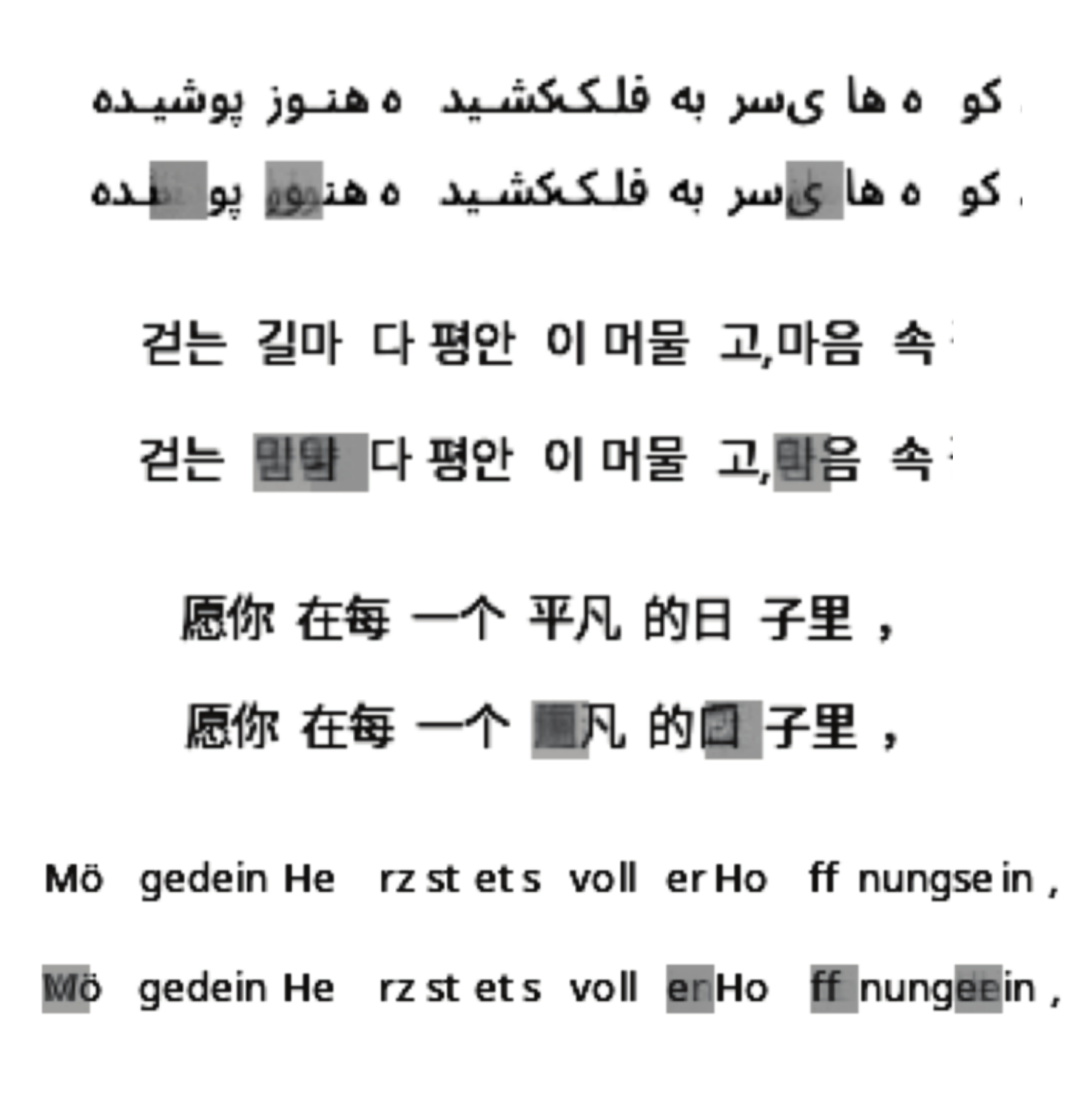}
    \caption{Examples of visual prediction from PLMs across multiple languages. The grayed areas are the PLM predictions for the regions originally masked.}
    \label{fig:prediction_examples}
\end{figure}

\subsection{Visual Word Prediction Difficulty (VWPD) in Context}
\label{method:1}
PLMs can quantify the prediction difficulty (VWPD) of the next word in the pixel space.
Given a sentence $\bm w = [w_1,\cdots,w_n]$ and its image patches $\bm X = [\bm x_1, \cdots, \bm x_N]$, suppose a target token $w_t$ spans from $s_t$-th to $e_t$-th patches $[\bm x_{s_t}, \cdots, \bm x_{e_t}] \subset \bm X$ in the rendered image ($1\leq s_t\leq e_t \leq N$).
A token may span over several patches, and a straightforward way to compute the next-word prediction difficulty is:
\begin{align}
    \vec{I}(w_t|\bm w_{<t}) := \sum_{k=s_t}^{e_t} \vec{s}(\bm x_k| \bm X_{<k}) \;\;\mathrm{,}
    \label{eq:surprisal}
\end{align}
\noindent
where the target patches $[\bm x_{s_t}, \cdots, \bm x_{e_t}]$ are masked one-by-one to enforce the prediction (black masking parts in Figure~\ref{fig:masking}).
A function $s(\cdot)$ quantifies the reconstruction error of the actual visual token (also see Figure~\ref{fig:prediction_examples} for examples of predictions).
For example, one can compute the instance-wise MSE in the pixel space: 
\begin{align}
    \label{eq:s_uni}
    \vec{s}(\bm x_k| \bm X_{<k})=\left\| \vec{\tilde{\bm x}}_k - \bm x_k \right\|_2^2 \;\;\mathrm{,} \\
    \vec{\tilde{\bm x}}_k = \mathcal{F}_k(\bm X_{<k} \circ m(\bm x_k)) \;\;\mathrm{,} 
\end{align}
where $\mathcal{F}_k$ returns the $k$-th patch of the output, $\circ$ denotes the concatenation of patches, and $m(\cdot)$ is a masking function.
A high prediction error of patches  $\vec{s}(\bm x_k| \bm X_{<k})$ is aggregated into the prediction difficulty of the word in Eq.~\ref{eq:surprisal}; a high value of $\vec{I}(w_t|\bm w_{<t})$ is expected to be linked with longer reading time.
Another reconstruction error is also examined in Section~\ref{sec:exp}.

Off-the-shelf PLMs, however, are typically implemented with a bidirectional encoder or encoder-decoder architecture, and fully omitting future context $\bm X_{>t}$ from the input may incur a severe training-inference gap.
Reflecting our vision, the future context is not fully invisible, but parafoveal sight still catches some superficial cues, e.g., approximate word length or sentence length.
Thus, we propose the simulation of parafoveal sight by feeding blurred future context information  $\bm X_{> t}$ with a blurring function $g(\cdot)$ (see Figure~\ref{fig:masking}).\footnote{
For brevity, $g(\cdot): \mathbb{R}^{P^2\times C}\to \mathbb{R}^{P^2\times C}$ can be applied element-wise if the argument is a sequence of patches, e.g.,  $g(\bm X)$ indicates $[g(\bm x_1), \cdots, g(\bm x_N)]$.}
This blurring design is ablated in the main experiments.
Now, the prediction difficulty of a word $w_t$ is redefined as follows:

\begin{align}
    &\overleftrightarrow{I}(w_t|\bm w_{\backslash t}) := \sum_{k=s_t}^{e_t} \overleftrightarrow{s}(\bm x_k| \bm X_{<k}, g(\bm X_{>k})) \;\;\mathrm{,}
    \label{eq:bi_surprisal} \\
        \label{eq:bidirectional_l2}
    &\overleftrightarrow{s}(\bm x_k| \bm X_{<k}, g(\bm X_{>k}))=\left\| \overleftrightarrow{\tilde{\bm x}}_k - \bm x_k \right\|_2^2 \;\;\mathrm{,} \\
    &\overleftrightarrow{\tilde{\bm x}}_k = \mathcal{F}_k(\bm X_{<k} \circ m(\bm x_k) \circ g(\bm X_{>k})) \;\;\mathrm{.} 
\end{align}

\noindent
where the prediction difficulty $\overleftrightarrow{I}(w_t|\bm w_{\backslash t})$ is conditioned by bi-directional context (with partial blurring to the future context).
Through our experiments, we use this version (Eq.~\ref{eq:bi_surprisal}) of visual word predictability.

\subsection{SSIM Reconstruction Measure}
\label{app:ssim}
For the reconstruction error term, $\vec{s}(\cdot)$ in Eq.~\ref{eq:s_uni} or $\overleftrightarrow{s}(\cdot)$ in Eq.~\ref{eq:bidirectional_l2}, another formulation beyond instance-wise MSE can also be examined.
For example, structural similarity (SSIM)~\cite{wang2004image} is widely used in the computer vision field.
Given that the SSIM is a similarity measure, one minus the SSIM can be adopted for estimating VWPD:

 \begin{align}
 \nonumber
&1- \mathrm{SSIM}(\tilde{\bm x}_k, \bm x_k) \\
&=1- 
\frac{(2\mu_{\tilde{x}}\mu_x + C_1)(2\sigma_{\tilde{x}x}+C_2)}
{(\mu_{\tilde{x}}^2+\mu_x^2+C_1)(\sigma_{\tilde{x}}^2+\sigma_x^2+C_2)}
\;\;\mathrm{,}
\end{align}
\noindent
where $\mu_{\tilde{x}}$ and $\mu_x$ denote the mean pixel intensities of the reconstructed patch $\tilde{\bm x}_k$ and ground-truth patch $\bm x_k$, respectively.
$\sigma_{\tilde{x}}^2$ and $\sigma_x^2$ denote their variances,
$\sigma_{\tilde{x}x}$ denotes the covariance between them,
and $C_1$ and $C_2$ are small constants for numerical stability.
One can apply this formula in Eq.~\ref{eq:s_uni} or Eq.~\ref{eq:bidirectional_l2} instead of instance-wise MSE, and we preliminarily examine this variant in Section~\ref{subsec:pre_exp}.

\subsection{Visual Prediction Uncertainty in Context}
\label{app:uncertainty}
We also examined the entropy of next-word prediction in the pixel space, which quantifies the uncertainty in predicting black (stroke) or white (no stroke) rather than noisy gray predictions.
Each dimension in a reconstructed pixel $\hat{\bm p}_k$ at position $k$ is first normalized with a sigmoid function (0 for white pixel prediction and 1 for black pixel prediction):

\begin{align}
\hat{\bm p}_k=\sigma(\overleftrightarrow{\hat{\bm{x}}}_k) \in [0,1]^{P^2\times C}\;\;\mathrm{.}
\end{align}
\noindent
The entropy measure is computed as follows:

\begin{align}
\label{eq:entropy}
H(w_t|\bm w_{\backslash t}) &:= \sum_{k=s_t}^{e_t} \sum_{l=1}^P\sum_{m=1}^P\sum_{c=1}^C h(\hat p_{k,l,m,c}) \;\;\mathrm{,}\\
h(p) &= -p \log p - (1-p)\log(1-p) \;\;\mathrm{,}
\end{align}
\noindent
where $\hat{p}_{k,l,m,c} \in [0,1]$ denotes ($l$, $m$)-th element in $c$-th channel of $k$-th patch prediction.
A high uncertainty value may parallel with the general difficulty of anticipating the next word, expected to be linked with longer reading time.
We also examine the entropy measures from PLMs and TLMs for word skipping modeling.

\section{Experimental Settings}
\label{sec:exp}

\subsection{Language Models}
We examine PLMs as well as TLMs with a comparable model size. All model details are in Table \ref{tab:checkpoints} in Appendix \ref{app:checkpoints}.

\paragraph{PLM Settings.}
For English experiments, we evaluate \texttt{pixel-m4}. For Chinese reading, we also use the off-the-shelf \texttt{pixel-m4} model \cite{kesen2025multilingual}, which has been pretrained on Chinese script. For other languages, we finetune \texttt{pixel-m4} on 100,000 examples from the \texttt{allenai/c4} dataset~\cite{raffel2020exploring} for 150,000 steps with a batch size of \texttt{32}. The finetuned models reach a loss below 0.020, comparable to the official checkpoint on the pre-training data.
For the blurring function $g(\cdot)$ in Eq.~\ref{eq:bidirectional_l2}, we use Gaussian Blurring with a kernel size of 5 and a sigma of 3 (see Appendix~\ref{app:noise}).
We also examine a no-blurring condition, in which the future context is fully visible, similarly to BERT~\cite{devlin2019bert}.

\paragraph{TLM Settings.}
As baselines, we use monolingual GPT2~\cite{radford2019language}, monolingual BERT~\cite{devlin2019bert}, and multilingual mGPT \cite{shliazhko2024mgpt}. 
Comparisons between TLMs and PLMs with parafoveal information clarify the general advantage of orthographically grounded prediction measures compared to discrete token prediction without any access to future word information.
BERT is also compared, given the bi-directional nature of the PLM encoder.
We use the surprisal \cite{hale2001, levy2008expectation} and entropy~\cite{hale2016information} computed by TLMs as predictors. 
Specifically, for the GPT and mGPT models, the surprisal of a word is calculated as:
\begin{equation}
    \vec{I}_\mathrm{tok}(w_t|\bm w_{<t}) = -\log_2 p(w_t \mid w_1, \ldots, w_{t-1}) \;\;\mathrm{.}
\end{equation}

\noindent
For the BERT models, the surprisal is computed with a bidirectional context:

\begin{align}
    \overleftrightarrow{I}_\mathrm{tok}(w_t|\bm w_{\backslash t}) &= -\log_2 p(w_t \mid w_1, \ldots, w_{t-1}, \notag \\
    &\quad \texttt{[MASK]}, w_{t+1}, \ldots, w_n) \;\;\mathrm{.}
\end{align}
\noindent
Entropy is calculated as the expected value of surprisal over the model's tokenizer vocabulary~\cite{pimentel2023effect}.

\begin{table}[t]
\centering
\scriptsize
\tabcolsep=0.07cm
\begin{tabular}{llllrcl}
\toprule
\textbf{Lang.}                    & \textbf{Dataset Code}                       & \textbf{Data}                & \makecell{\textbf{Skip}\\\textbf{Rate}} & \makecell{\textbf{Avg.}\\\textbf{TRT}} & \textbf{\#Subj.} & \textbf{Script}                 \\
\cmidrule(r){1-1} \cmidrule(lr){2-3} \cmidrule(lr){4-5}  \cmidrule(lr){6-6} \cmidrule(l){7-7}
\multirow{2}{*}{German}     & Deu (meco po) & MECO-2 (ge\_po)     & 0.19          & 380.8       & 40 & \multirow{7}{*}{Latin} \\
                            & Deu (meco zu) & MECO-2 (ge\_zu)     & 0.21          & 390.6       & 45 &                        \\
\cmidrule(r){1-1} \cmidrule(lr){2-3} \cmidrule(lr){4-5}  \cmidrule(lr){6-6}
\multirow{2}{*}{English}    & Eng (meco en) & MECO-1 (en)         & 0.30          & 346.1       & 46 &                        \\
                            & Eng (meco uk) & MECO-2 (en\_uk)     & 0.28          & 348.2       & 50 &                        \\ 
\cmidrule(r){1-1} \cmidrule(lr){2-3} \cmidrule(lr){4-5}  \cmidrule(lr){6-6}
\multirow{2}{*}{Spanish}    & Esp (meco es) & MECO-1 (es)         & 0.33          & 406.7       & 48 &                        \\
                            & Esp (meco ch) & MECO-1 (es\_ch)     & 0.29          & 367.7       & 45 &                        \\ 
\cmidrule(r){1-1} \cmidrule(lr){2-3} \cmidrule(lr){4-5}  \cmidrule(lr){6-6} \cmidrule(l){7-7}
Persian                     & Far           & PSR                 & 0.13          & 450.4       & 60 & \multirow{1}{*}{Abjad} \\ 
\cmidrule(r){1-1} \cmidrule(lr){2-3} \cmidrule(lr){4-5}  \cmidrule(lr){6-6} \cmidrule(l){7-7}
Korean                      & Kor           & MECO-1 (ko)         & 0.31          & 348.9       & 32 & Hangul                 \\
\cmidrule(r){1-1} \cmidrule(lr){2-3} \cmidrule(lr){4-5}  \cmidrule(lr){6-6} \cmidrule(l){7-7}
\multirow{2}{*}{Chinese}    & Simp (meco)   & MECO-2 (ch-s)       & 0.43          & 348.7       & 39 & \multirow{2}{*}{Hanzi} \\
 \cmidrule(lr){2-3} \cmidrule(lr){4-5}  \cmidrule(lr){6-6}
                            & Trad (meco)   & MECO-2 (ch-t)       & 0.44          & 435.3       & 38 &                        \\ 
\cmidrule(r){1-1} \cmidrule(lr){2-3} \cmidrule(lr){4-5}  \cmidrule(lr){6-6} \cmidrule(l){7-7}
Turkish                     & Tur           & MECO-2 (tur)        & 0.19          & 363.7       & 16 & Latin                 \\
\bottomrule
\end{tabular}
\caption{We use multilingual and multi-script eye-movement-while-reading data across 459 readers.}
\label{tab:Statistics}
\end{table}

\subsection{Human Reading Data}
We model reading time and word skipping behavior of a total of 459 subjects from different eye-movement-while-reading corpora, covering German (Deu), English (Eng), Spanish (Esp), Farsi (Far), Korean (Kor), Simplified Chinese (Simp), Traditional Chinese (Trad), and Turkish (Tur). Data are from MECO Wave 1 \cite{siegelman2022expanding} or Wave 2~\cite{siegelman2025wave}, except for Persian reading corpus \cite{tekbudak2024psr}. Data statistics and details are presented in Table~\ref{tab:Statistics}.
We take the following reading measures into consideration: First Fixation Duration (FFD), First-pass Reading Time (FPT), Total Reading Time (TRT), First-pass Skipping (FSKP), and Overall Skipping (SKP). For durative measures, any zero fixation duration and zero reading time are excluded. 
We adopt two word skipping measures: (i) first-pass skipping --- whether an interest area is skipped before a later interest area is fixated (1=yes, 0=no); and (ii) Overall skipping --- whether an interest area is not fixated throughout the current trial (1=yes, 0=no).

\subsection{Regression Models}
\paragraph{Predictors of Interest.}
We extract visual prediction difficulty (Eq.~\ref{eq:bi_surprisal}) and visual entropy (Eq.~\ref{eq:entropy}) from PLMs, as well as the widely used token-level surprisal and entropy from TLMs. 
For each regression model, we only use one predictor of interest, denoted as $\mathtt{x}$.
The regression formula is as follows:
    \begin{align}
    \nonumber
    \{\mathtt{y_{n}}\} \sim\ & \{\mathtt{x_{n}}\} + \mathrm{l_{n}} + \mathrm{f}_{n} 
            + \mathrm{l_{n-1}} + \mathrm{f_{n-1}}  + \mathrm{idx_{n}} \\
            \nonumber
            & +  (1+\{\mathtt{x_{n}}\} +\mathrm{l_{n-1}} + \mathrm{f_{n-1}}|\mathrm{subject\_id}) \\ 
            & + (1|\mathrm{stimuli\_id}) \;\;\mathrm{,}
    \end{align}

where $\mathrm{l}$ is word length, and $\mathrm{f}$ is zipf word frequency, $\mathrm{idx_{n}}$ is the word index in sentence. The regression formula for the baseline is as follows:

    \begin{align}
    \nonumber
    \{\mathtt{y_{n}}\} \sim\ & \mathrm{l_{n}} + \mathrm{f}_{n} 
            + \mathrm{l_{n-1}} + \mathrm{f_{n-1}}  + \mathrm{idx_{n}} \\ 
            \nonumber
            & +  (1+\mathrm{l_{n-1}} + \mathrm{f_{n-1}}|\mathrm{subject\_id}) \\ 
            & + (1|\mathrm{stimuli\_id}) \;\;\mathrm{.}
    \end{align} 
Here, $\mathtt{y}$ is the reading measures, i.e., FFD, FPT, TRT, FSKP, or SKP.
For reading measures (FFD, FPT, and TRT), we applied log transformation, and skipping measures (FSKP and SKP) are converted with binomial family.
We conduct our main experiments with generalized additive mixed models (GAMM) \cite{soskuthy2017generalised, 10.1111/rssc.12068}, using \texttt{mgcv} packages \cite{simon} and setting all effects as linear. 
For main experiments, we use the following designs of $\mathrm{x}$:
\begin{itemize}
    \item $\mathrm{pixel}$: VWPD from PLM, without blurring the future context.
    \item $\mathrm{pixel\_uni}$: VWPD from PLM, with blurring the future context.
    \item $\mathrm{gpt}$: surprisal from the respective monolingual GPT2. 
    \item $\mathrm{mgpt}$: surprisal from multilingual GPT.
    \item $\mathrm{bert}$: bi-directional surprisal from the respective monolingual BERT.
    \item $\mathrm{pixel\_uni+gpt}$: $\mathrm{pixel\_uni}$ and $\mathrm{gpt}$ are both included, without interaction.
    \item $\mathrm{pixel\_uni*gpt}$: Interaction between $\mathrm{pixel\_uni}$ and $\mathrm{gpt}$ with their main effects.
\end{itemize}

\subsection{Evaluation Metrics}
Following existing studies~\cite{goodkind2018predictive}, we use the difference in loglikelihood between two nested models with and without the predictor of interest $\mathtt{x_{n}}$ from TLMs or PLMs. 
In our case, the joint contribution of the predictor and its subject-specific random slope is measured.
We refer to this metric as psychometric predictive power (PPP).
To test robustness, we rerun each formula 100 times using a crossed random-effects bootstrap with replacement to obtain a distribution of the log-likelihood difference. 
Similarly to~\citet{goodkind2018predictive}, we fit regression models to explain the data of interest, but in Appendix~\ref{app:cv}, we also adopt 10-fold cross-validation, as in \citep{wilcox2023testing}, yielding consistent results.
When reporting a single PPP value for each language with several datasets (Table~\ref{tab:Statistics}), PPPs for different datasets in the target language are averaged.

\begin{figure}[t]
    \centering
    \includegraphics[width=0.98\linewidth]{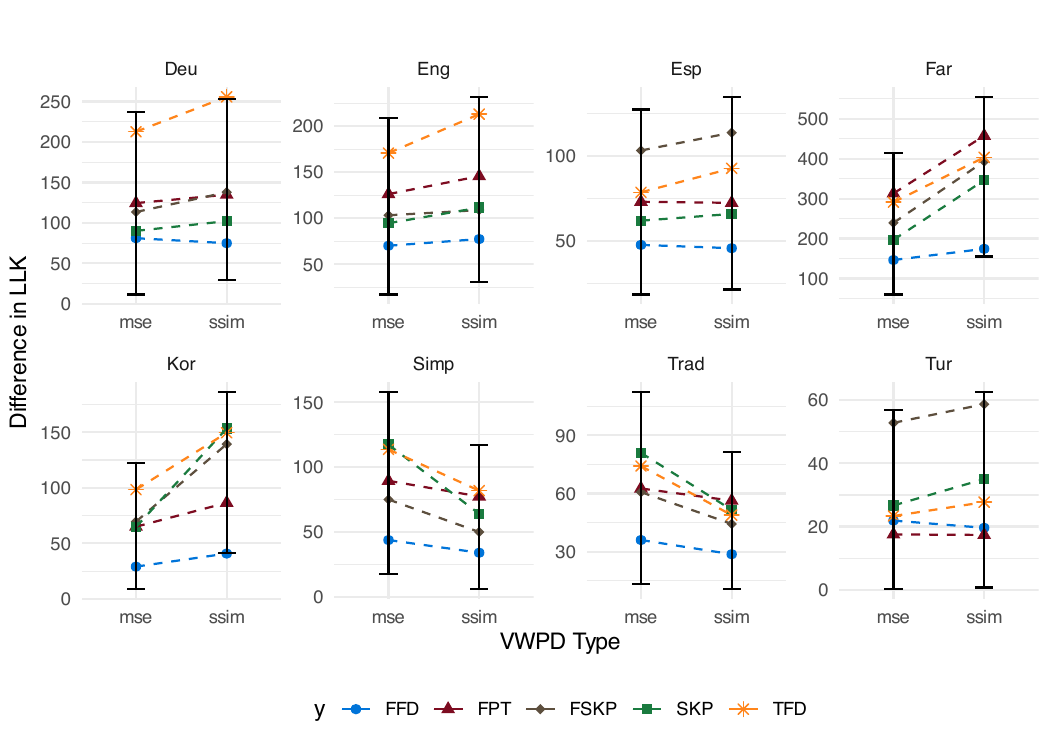}
    \caption{MSE-based and SSIM-based VWPD estimates show slight, but different PPP patterns across languages. The error bars in the figures indicate 95\% percentile bootstrap confidence intervals of PPP across the settings.}
    \label{fig:vwpcomparison}
\end{figure}

\begin{figure*}[!t]
    \centering
    \includegraphics[width=0.98\linewidth]{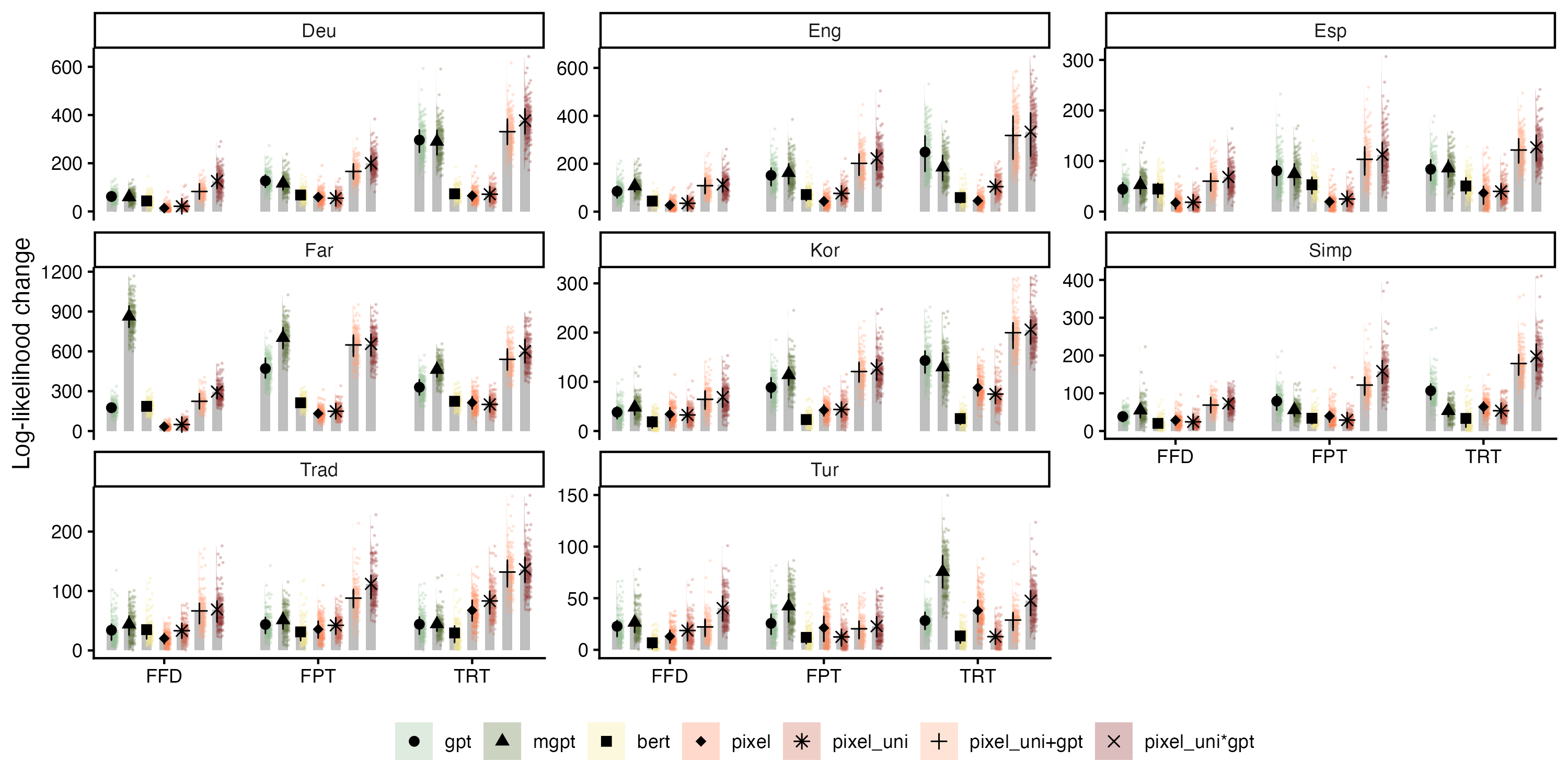}
    \caption{PPP (larger better) of VWPD and token surprisal for fixation measures (FFD = First Fixation Duration, FPT = First-pass Reading time, TRT = Total Reading Time). All zero-duration or zero-reading-time is filtered to disentangle from skipping measures. The Y-axis is PPP. The error bars in the figures indicate standard deviation.}
    \label{fig:plot_surprisal_rt}
\end{figure*}

\begin{figure*}[!t]
    \centering
    \includegraphics[width=0.98\linewidth]{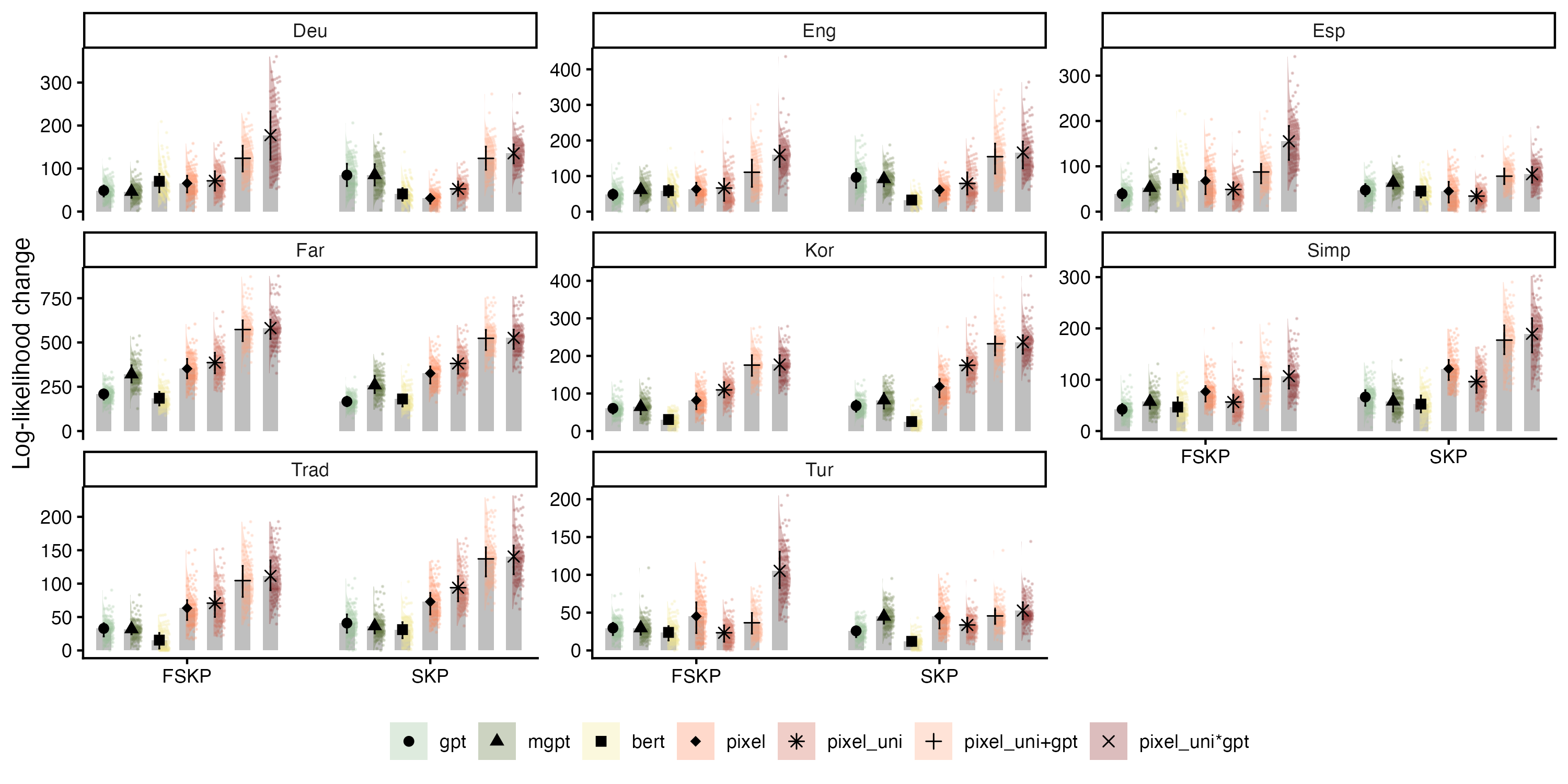}
    \caption{PPP (larger better) of VWPD and token surprisal for skipping measures (FSKP = First-pass skipping, SKP = Overall skipping). The Y-axis is PPP. The error bars in the figures indicate standard deviation.} %
    \label{fig:plot_surprisal_skp}
\end{figure*}

\begin{figure*}[t]
    \centering
    \includegraphics[width=\linewidth]{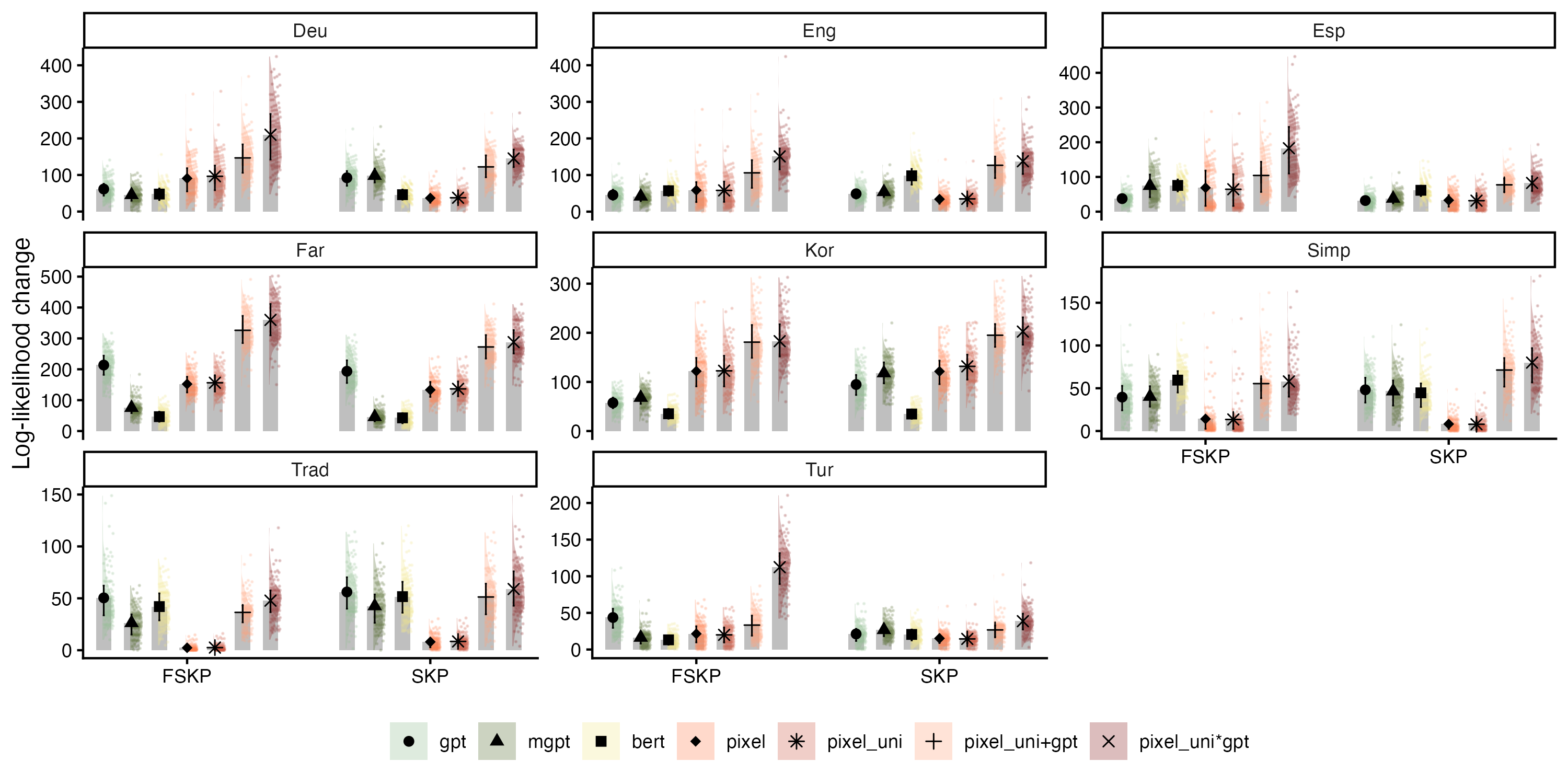}
    \caption{PPP (larger better) of visual and token entropy for skipping measures (FSKP = First-pass skipping, SKP = Overall skipping). The Y-axis is PPP. The error bars in the figures indicate standard deviation.}
    \label{fig:entropy}
\end{figure*}

\section{Experimental Results}
\label{sec:results}
\subsection{Preliminary Experiments: Comparing Two VWPD Measures}
\label{subsec:pre_exp}
In Section~\ref{method:1}, we introduced
 MSE and SSIM as an image reconstruction error in Eq.~\ref{eq:bidirectional_l2}.
As preliminary experiments, we compare MSE-based and SSIM-based image reconstruction errors as general indicators of next-word prediction difficulty.
Figure~\ref{fig:vwpcomparison} shows that different PPPs are obtained with different error formulations, and there was a clear empirical trend that SSIM is slightly better or comparable to MSE, except for the Chinese (Simplified and Traditional) language.
This might reflect a special visual processing of logograms or the complexity of Chinese characters.
In our main experiments, we show SSIM results across all languages, except for MSE for the Chinese (Simplified and Traditional) data.
Full SSIM and MSE comparison is shown in Appendix~\ref{app:mse_ssim}.

\subsection{Comparing PLMs and TLMs}
We evaluate two versions of PLMs with and without blurring for the future context, and three TLMs (monolingual GPT2, BERT, and mGPT).

\paragraph{Fixation and Reading Time} 
Figure~\ref{fig:plot_surprisal_rt} shows the results for FFD, FPT, and TRT modeling.
Visual word predictability (VWPD) is used for PLMs, and surprisal is used for TLMs.
PPPs for PLMs are somewhat worse than TLMs, but when these two predictors are jointly used, the PPP substantially increases, indicating their complementary advantage in reading time modeling.

\paragraph{Word Skipping} 
Figure \ref{fig:plot_surprisal_skp} shows the results for FSKP and SKP; whether the word is skipped or not during reading is predicted.
In contrast to gaze duration measures, PLMs yield comparable or better predictive power, especially for the Farsi (Far), Korean (Kor), and Chinese (Simp and Trad) languages; interestingly, these languages do not adopt the Latin alphabet.
In these languages, in other words, if a word can easily be predicted visually from the surrounding context, it is likely to be skipped during reading. 
This idea is also compatible with entropy measures (how the next word is uncertain), and Figure~\ref{fig:entropy} shows the comparison of entropy computed by PLMs and TLMs.
The results generally reproduce our observations with surprisal, i.e., token-level and visual entropy are complementary.
We report the cross-validation analysis in appendix \ref{app:cv}, which substantiates the tendencies we have pointed out above.

\subsection{Comparing Addition and Interaction}
Comparing the PPP of the additive model ($\mathrm{pixel\_uni\\+gpt}$) and the interactive model ($\mathrm{pixel\_uni*gpt}$), we sometimes observe an advantage of the interaction term, especially in the first-pass measures (FFD and FSKP) in German (Deu), Spanish (Esp), and Turkish (Tur).
These have somewhat longer word length and richer morphology than those of other languages~\cite{siegelman2022expanding}. 
Thus, a strong interaction may imply a more general hierarchical engagement of prediction in reading, spanning from shallow visual perception to morphological or richer linguistic processing.
In the following section, we look into how this interaction shapes first-pass reading behavior. 
The detailed breakdown of our main results into individual dataset is in Figures~\ref{fig:detailresultreadingtime} and \ref{fig:detailresultskiping}.

\section{Analysis: Characteristics of VWPD}

\subsection{Interaction with TLM Surprisal}Given the complementary advantage of predictors from PLM and TLM, we exploratorily extend the analysis, specifically on how PLM and TLM predictors interact. We focus on early-stage reading measures, such as first fixation duration (FFD), which typically aligns with early lexical access~\cite{kurnik2016bilingual}, first-pass reading time (FPT), as well as first-pass refixation duration (FPT minus FFD; FPREF), which indicates the ``failure to achieve lexical access''~\cite{lowell2014word}. For simplicity, we only consider the fixed effect of PLM surprisal (namely VWPD) and TLM surprisal (we use GPT2 surprisal in each language).
We split the data into those with relatively high surprisal (+1 SD) and low surprisal ($-$1 SD) and analyze how the VWPD effect differs between the two subsets.

\begin{figure*}[t]
    \centering
    \includegraphics[width=0.98\linewidth]{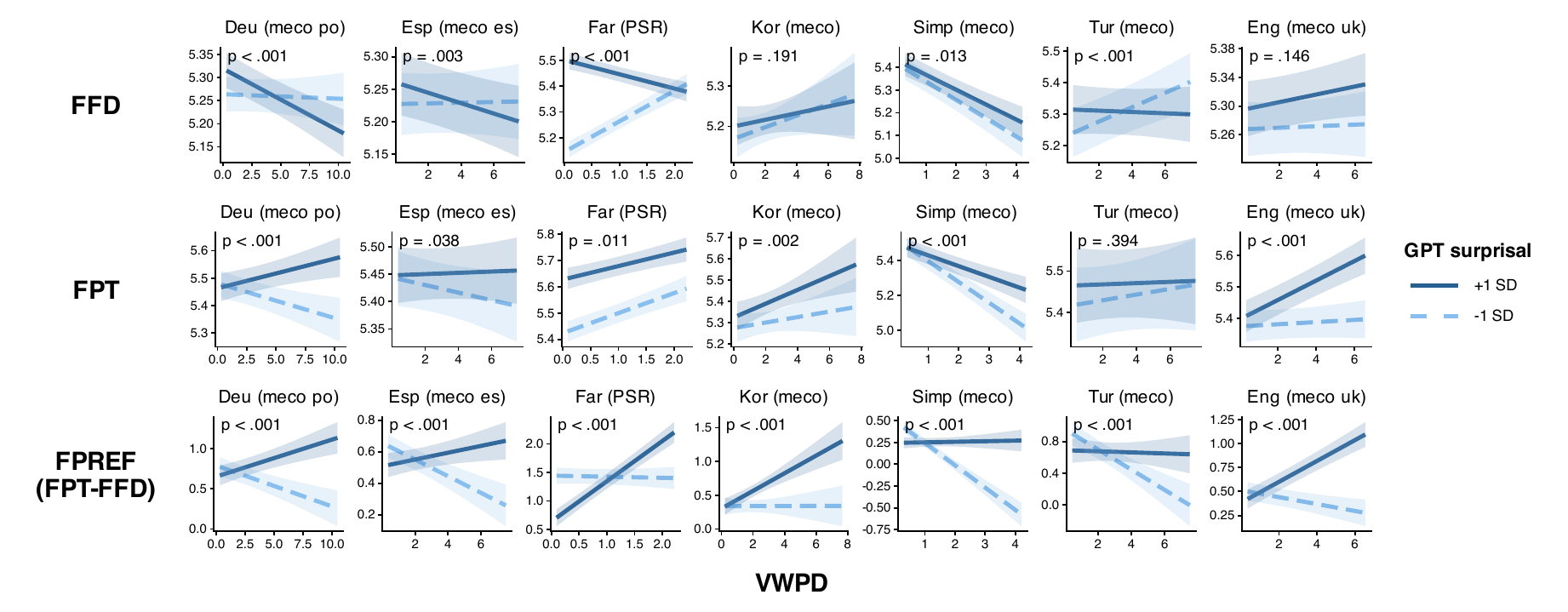}
    \caption{Interaction between VWPD and GPT2 surprisal (+1 SD = high, $-$1 SD = low) to FFD, FPT, and FPREF. 
    }
    \label{fig:interact}
\end{figure*}

Figure~\ref{fig:interact} shows the interaction between VWPD and GPT2 surprisal. 
Interestingly, for FFD, we found that high surprisal (+1 SD in the figure) can reverse (e.g., in Deu and Tur) the VWPD--FFD relationship to be negative. 
 In contrast to FFD, FPREF results indicate a positive mediation of VWPD, as high surprisal (+1SD in the figure) leads to a positive VWPD-FPREF relationship. 
That is, words with a high surprisal (lexical difficulty) tend to weaken the VWPD effect for first fixation but strengthen it for refixation. 
One possible interpretation of the neutralized or negative interaction between VWPD and surprisal in FFD prediction is that lexical prediction (token-level surprisal) may involve pre-activation of visual representations \cite{buggy2026predictability}; if more effort is devoted to predicting the next word, early-stage visual processing can be skipped.
The positive interaction in FPREF aligns with the finding that orthographic confusion is significant only for unpredictable words \cite{yao2022predictability}, as the positive interaction was only observed when people need multiple fixations (FPREF), potentially when lexical access initially failed.

\begin{figure}[t]
    \centering
    \includegraphics[width=\linewidth]{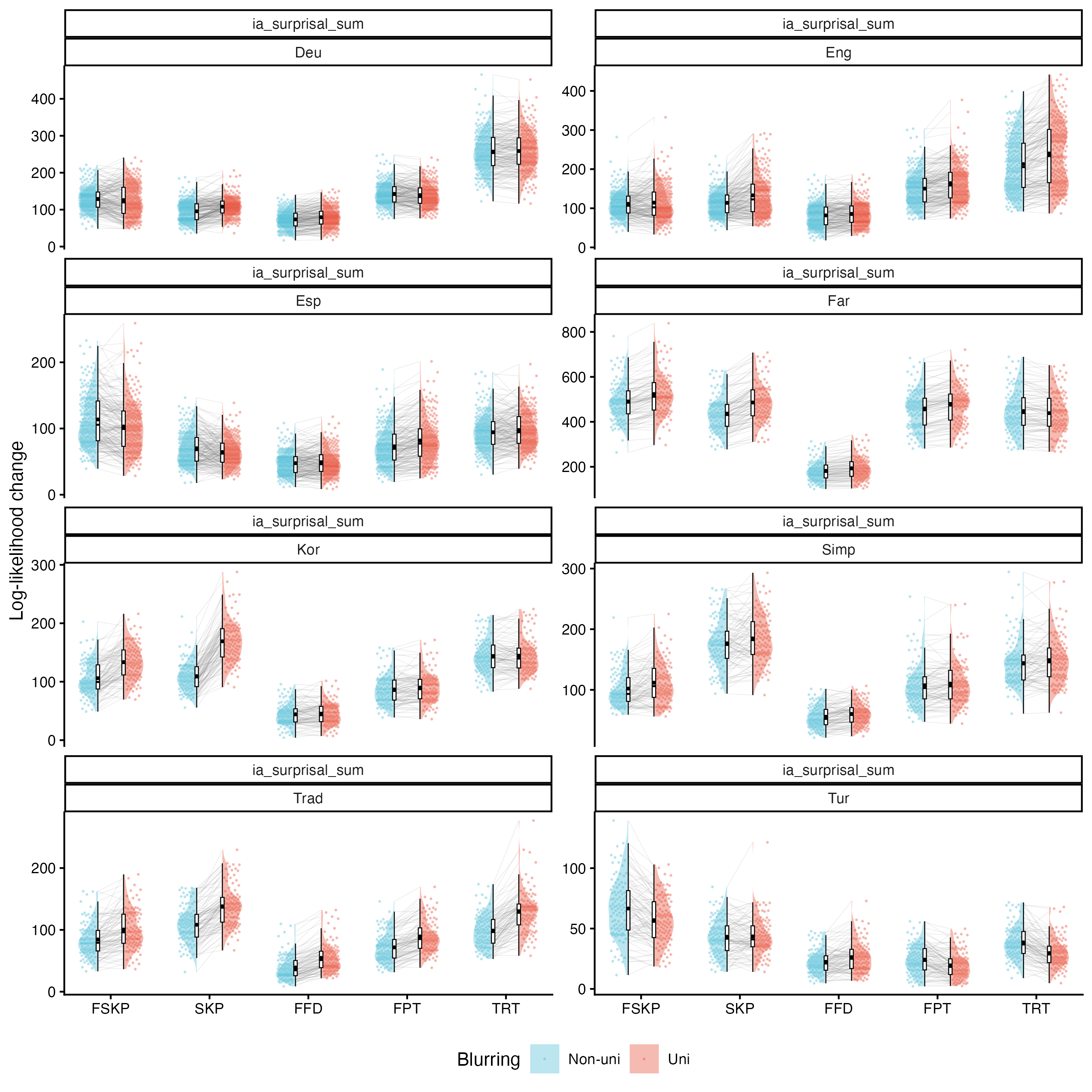}
    \caption{Effect of Unidirectional Gaussian Blurring. The points are obtained through bootstrapping, and the blue (left) ones are from the \texttt{pixel} setting without blurring, while the red (right) ones are from the \texttt{pixel\_uni} one with blurring. Y-axis is PPP.}
    \label{fig:effectofuni}
\end{figure}

\subsection{Gaussian Blurring Effects}
We also analyze how Gaussian blurring of the future context affects PPP. 
The improvement in predictive power (See Figure \ref{fig:effectofuni}) with the blurring treatment is not substantial. 
Still, the advantage is slightly pronounced for SKP in Farsi (Far), Korean (Kor), and Traditional Chinese (Trad), where non-Latin alphabets are used.
The advantage of blurring in Trad has also been suggested by the finding that reading Traditional Chinese characters narrows the perceptual span \cite{pan2024perceptual}, supporting an information-processing account of visual perception during reading, as assumed in the OB1-Reader model~\cite{snell2018ob1}.
Nevertheless, our Gaussian Blurring is our tentative implementation, and more cognitively motivated designs should be explored in future studies.

\section{Conclusion}
We demonstrate that pixel-based LMs can estimate expectation-based visual word-processing measures that complement those estimated by token-based LMs, which are especially relevant to pre-fixation processing on partial input. 
This study is the first to introduce a pixel-based language model into computational cognitive modeling. The results demonstrate, at least partial cognitive plausibility of PLM, especially its alignment with early-stage language processing, such as first-pass skipping. %

\section*{Limitations}
First, a pixel-based language model has various rendering strategies \cite{lotz2023text}. We only use the most cross-lingually generalizable setting, namely bigram rendering, which is the foundation of the only multilingual pixel-based model so far. We agree that calculating visual word prediction metrics is highly sensitive to the text rendering strategy. We encourage future work to explore different rendering strategies and analyze their effect on the cognitive plausibility of pixel-based language modeling.
Second, for pixel blurring, we have not conducted an intensive hyperparameter analysis regarding the effects of kernel size and sigma on model performance. We also only tried Gaussian Blurring, without exploring other image corruption methods such as Gaussian Noise. Future work could also compare more cognitively motivated parafoveal blurring and noise. Besides, the human parafovea exhibits graded visual acuity, which is not accounted for in this study. We hope that the current study initiates the primary work and establishes the foundational settings, and that future work will continue to explore parafoveal vision modeling by manipulating the gradedness of visual acuity.

\section*{Ethical Considerations}
This study uses open-source eye-movement-while-reading data collected from IRB-approved studies. No personally identifiable information is used in our experiment. Our study aims to gain knowledge about human cognition. We do not curate any system to replace humans in terms of reading. 

\section*{AI Writing/Coding Assistance Policy}
AI writing assistance systems like Grammarly and ChatGPT are occasionally used to correct grammatical and spelling errors and to adjust phrasing. 
Our use of writing assistants complies with the AI writing assistance policy described in ARR guidelines under category (a), assistance purely with the language of the paper. 

\section*{Acknowledgments}
This work was supported by JSPS KAKENHI Grant Numbers 23K16938 and 26H02511.

\bibliography{custom}

\clearpage
\appendix

\section{Models}
\label{app:checkpoints}
In Table \ref{tab:checkpoints}, we present the details of models used in this study.
\begin{table*}[t]
\centering
\small
\tabcolsep=0.05cm
\begin{tabular}{llrlr}
\toprule
Lang. & Unidirectional LMs & \makecell{Model size\\(\#parameters)} & Bidirectional LMs & \makecell{Model size\\(\#parameters)} \\ 
\cmidrule(r){1-1} \cmidrule(lr){2-3} \cmidrule(lr){4-5}
Deu & dbmdz/german-gpt2 & 510MB ($\sim$137M) & google-bert/bert-base-german-cased & 439MB ($\sim$109M) \\
Eng & openai-community/gpt2 & 548MB ($\sim$137M) & google-bert/bert-base-cased & 436MB ($\sim$108M) \\
Esp & DeepESP/gpt2-spanish & 548MB ($\sim$124M) & dccuchile/bert-base-spanish-wwm-cased & 440MB ($\sim$110M) \\
Far & gpt2-persian & 1.31GB ($\sim$328M) & HooshvareLab/bert-base-parsbert-uncased  & 654MB ($\sim$163M) \\
Kor & skt/kogpt2-base-v2 & 513MB  ($\sim$125M) & kykim/bert-kor-base & 476MB ($\sim$118M) \\
Simp (zh) & uer/gpt2-chinese-cluecorpussmall & 421MB ($\sim$102M) & google-bert/bert-base-chinese & 412MB ($\sim$102M) \\
Trad (zh) & ckiplab/gpt2-base-chinese & 421MB ($\sim$102M) & ckiplab/bert-base-chinese & 409MB ($\sim$102M) \\
Tur & ytu-ce-cosmos/turkish-gpt2 & 498MB ($\sim$124M) & dbmdz/bert-base-turkish-cased & 445MB ($\sim$111M) \\
\cmidrule(r){1-1} \cmidrule(lr){2-3} \cmidrule(lr){4-5}
All & mGPT & 3.45GB ($\sim$1.4B) & Pixel-m4 (including finetuned ones) & 449MB ($\sim$112M) \\
\bottomrule
\end{tabular}
\caption{Models with their sizes in our experiment. All checkpoints of GPT and BERT are from \url{https://huggingface.co/}.}
\label{tab:checkpoints}
\end{table*}
\section{PLM details}
\label{app:plm}

\subsection{Conditions}

\begin{figure*}
    \centering
    \includegraphics[width=\linewidth]{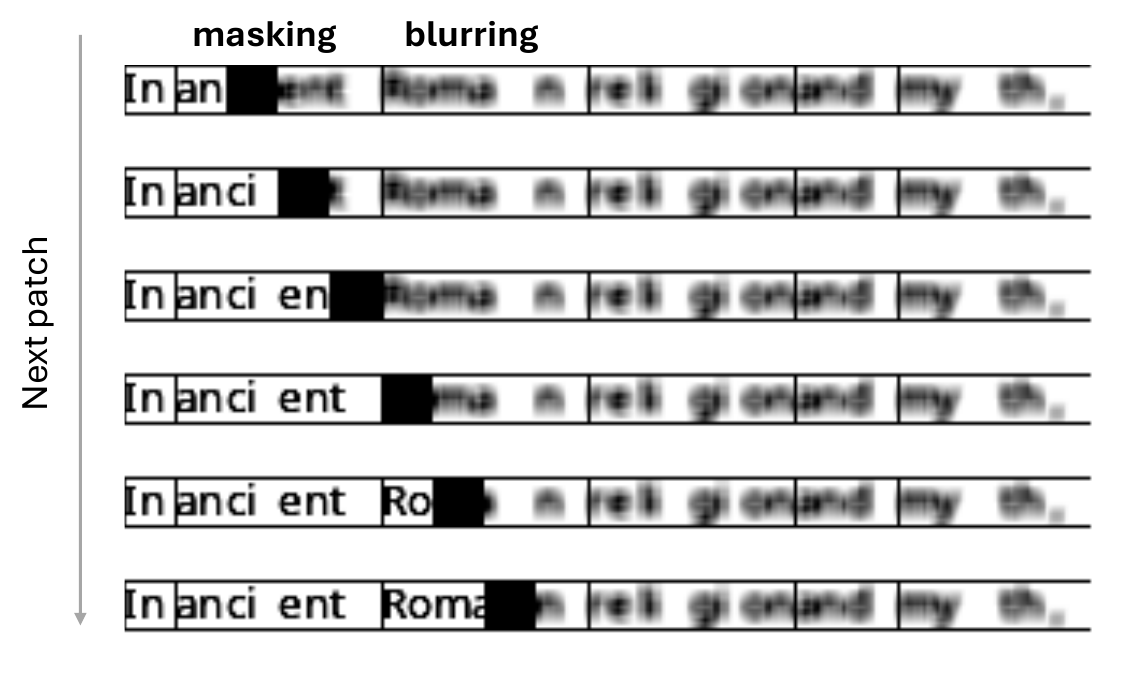}
    \caption{We approximate next-word predictability with patch-by-patch masking and add noise for the future context, simulating parafoveal sight, using \textsc{PLM}.}
    \label{fig:four_conditions}
\end{figure*}

Figure~\ref{fig:four_conditions} shows all four conditions explored in this study.
The first option is whether to mask all the patches related to the target word (whole masking) or only the target patch (partial masking) when computing patch predictability.
The second option is whether to apply the noisy blurring to the future context, potentially simulating the parafoveal information, or not.

\subsection{Gaussian Blurring}
We apply the following noise function to the original image $\bm X$:
\label{app:noise}
\begin{align}
    \bm{X}'(i, j) &= \sum_{k=-m}^{m} \sum_{l=-m}^{m} \bm{X}(i-k,\; j-l) \cdot G(k,\, l) \\[10pt]
    \quad m &= \left\lfloor \frac{\texttt{kernel\_size-1}}{2} \right\rfloor \\
    G(k, l) &= \frac{1}{2\pi\sigma^2} e^{-\frac{k^2 + l^2}{2\sigma^2}} \\[10pt]
\end{align}
As a cognitively implausible baseline, we also examine a no-blurring condition in which the right-side context is visible.

\subsection{Others}
To adapt Persian to the bigram renderer, we leave the space between two characters, even if they are connected visually. However, the space does not change the pre-space word into an unconnected form if it is originally in a connected form. To gather VWPD from Persian, we did not include the white-area. For the results in the current paper, we used the checkpoint after the 33,500th step results of the PSR dataset.

\section{Detailed Results}
\label{app:results}

\subsection{Interactive Effect of PLM prediction difficulty to TLM surprisal}
\begin{figure}[t]
    \centering
    \includegraphics[width=\linewidth]{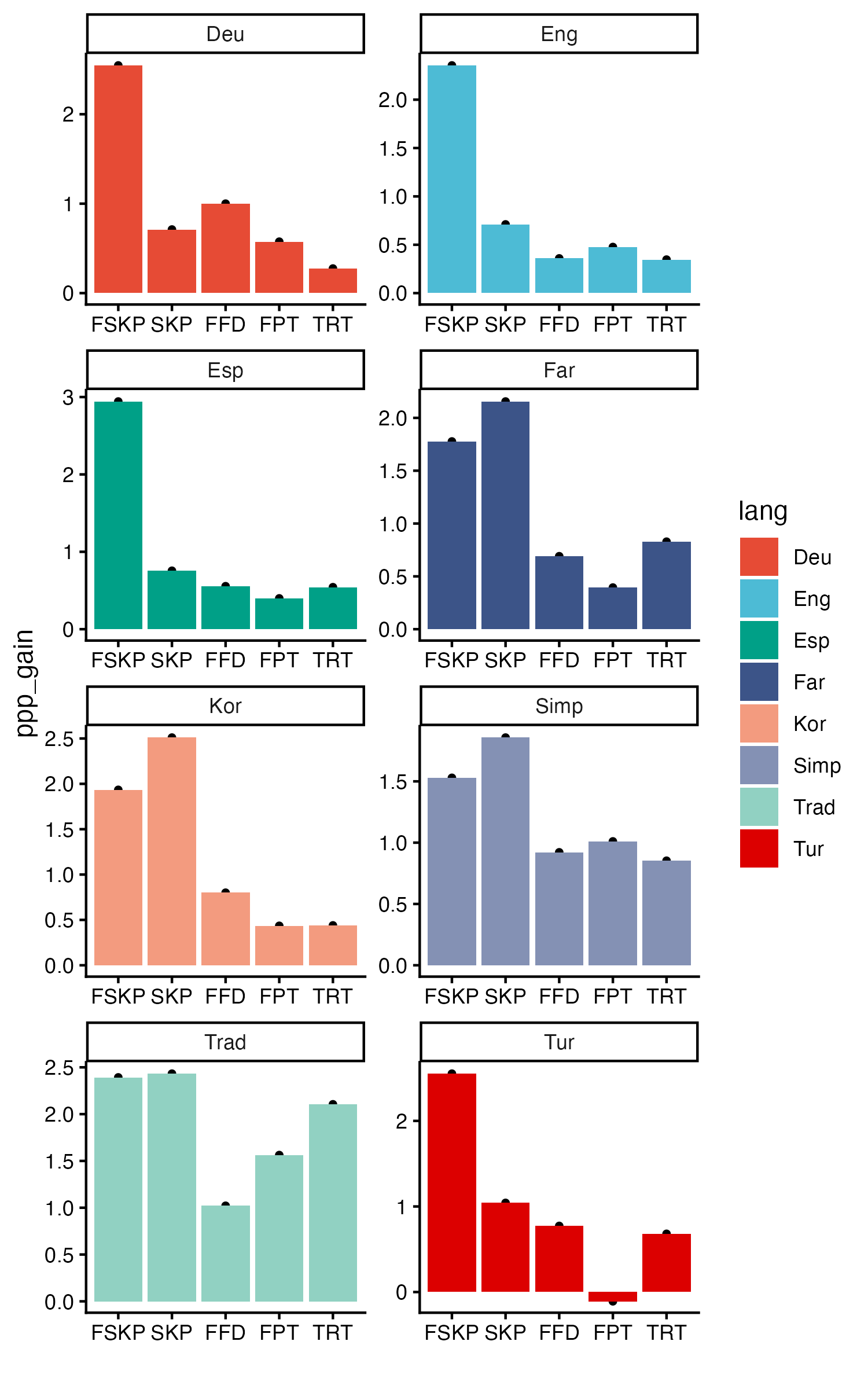}
    \caption{Gain of PPP after introducing PLM predictability to TLM ($\frac{\mathrm{PPP}_{\mathrm{pixel\_uni}*\mathrm{gpt}}-\mathrm{PPP}_{\mathrm{gpt}}}{\mathrm{PPP}_{\mathrm{gpt}}}$), which suggests multimodal prediction encodes rich metacognitive knowledge about skipping.}
    \label{fig:gain}
\end{figure}

Figure~\ref{fig:gain} shows a more detailed comparison between the regression model with GPT surprisal and that with GPT surprisal and the interaction between GPT surprisal and PLM VWPD.

\FloatBarrier

\subsection{Effects of Whole Word Masking}
\begin{figure}[t]
    \centering
    \includegraphics[width=\linewidth]{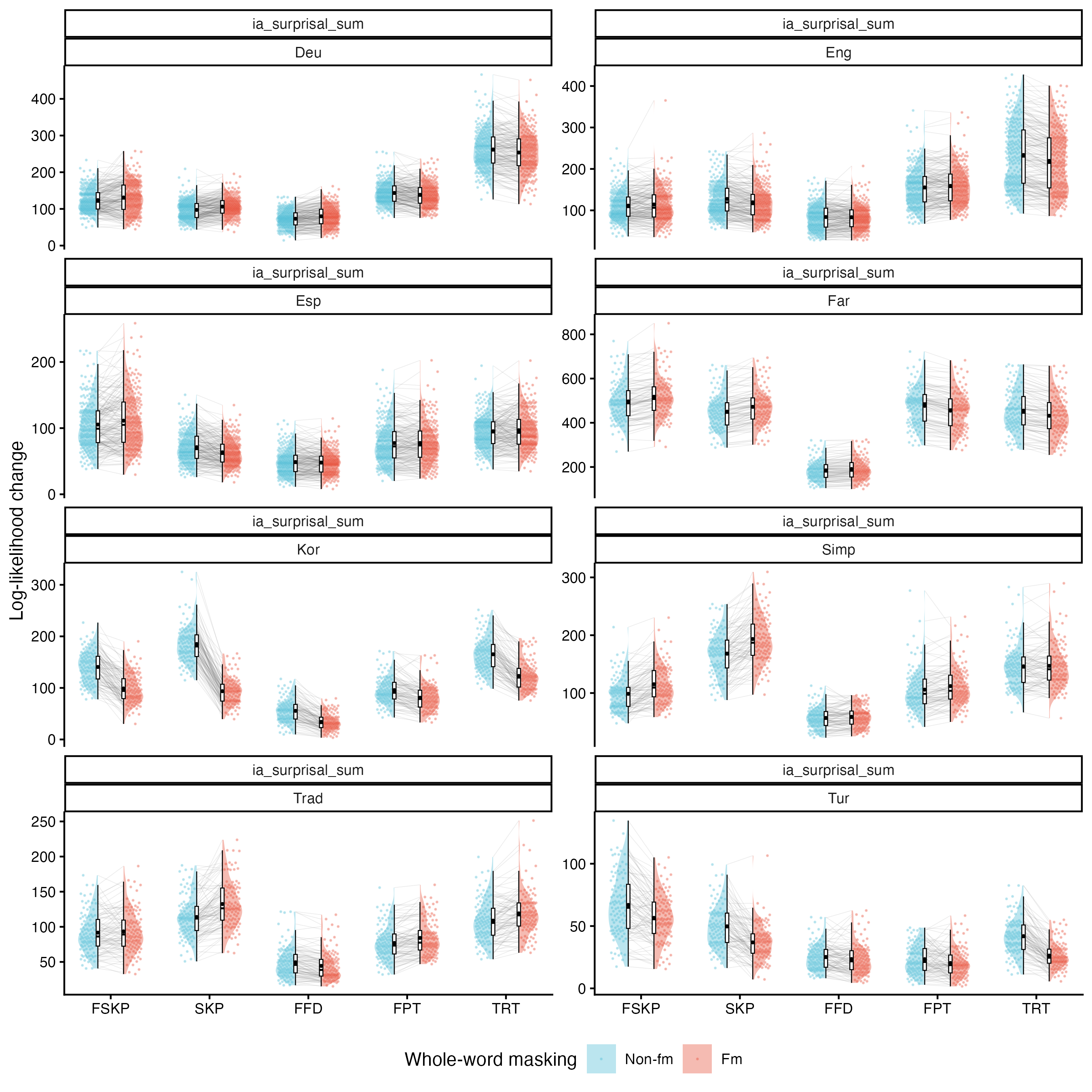}
    \caption{Effect of Whole Masking on PPP (Paired-t test of PPP is implemented, with matched regression model formula)}
    \label{fig:effectofmask}
\end{figure}
\FloatBarrier

Compared with blurring, the effect of masking is more nuanced across languages. We notice an advantage of whole-word masking in word skipping in Traditional Chinese, but a disadvantage in Korean. This potentially indicates a language-dependent sub-word processing strategy during naturalistic reading.

\subsection{Main Results by Dataset}
\label{app:detairesults}
Below are the detailed main results for each dataset:
\begin{figure*}[t]
    \centering
    \includegraphics[width=\linewidth]{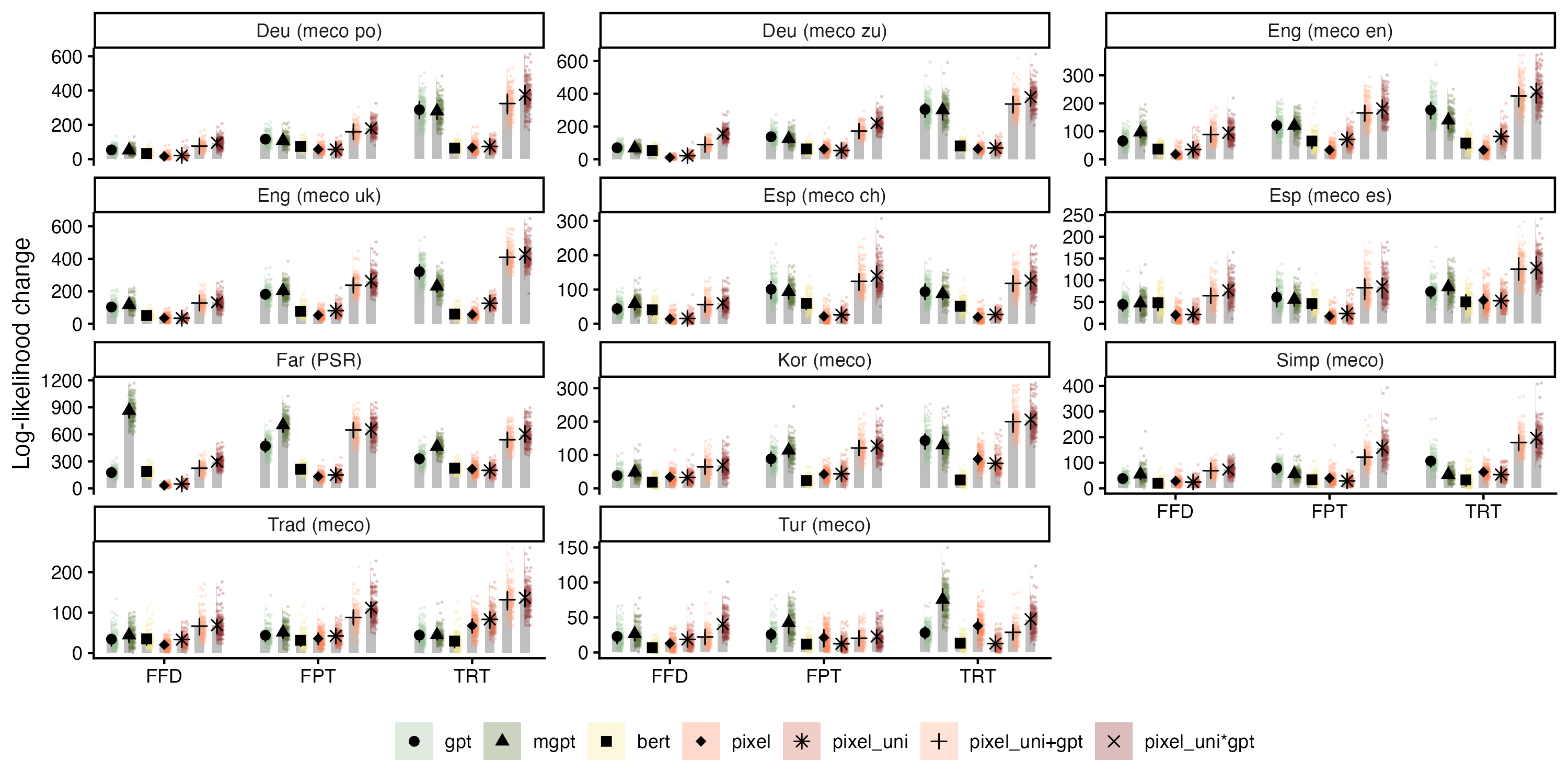}
    \caption{Main results by dataset (Reading Time)}
    \label{fig:detailresultreadingtime}
\end{figure*}
\begin{figure*}[t]
    \centering
    \includegraphics[width=\linewidth]{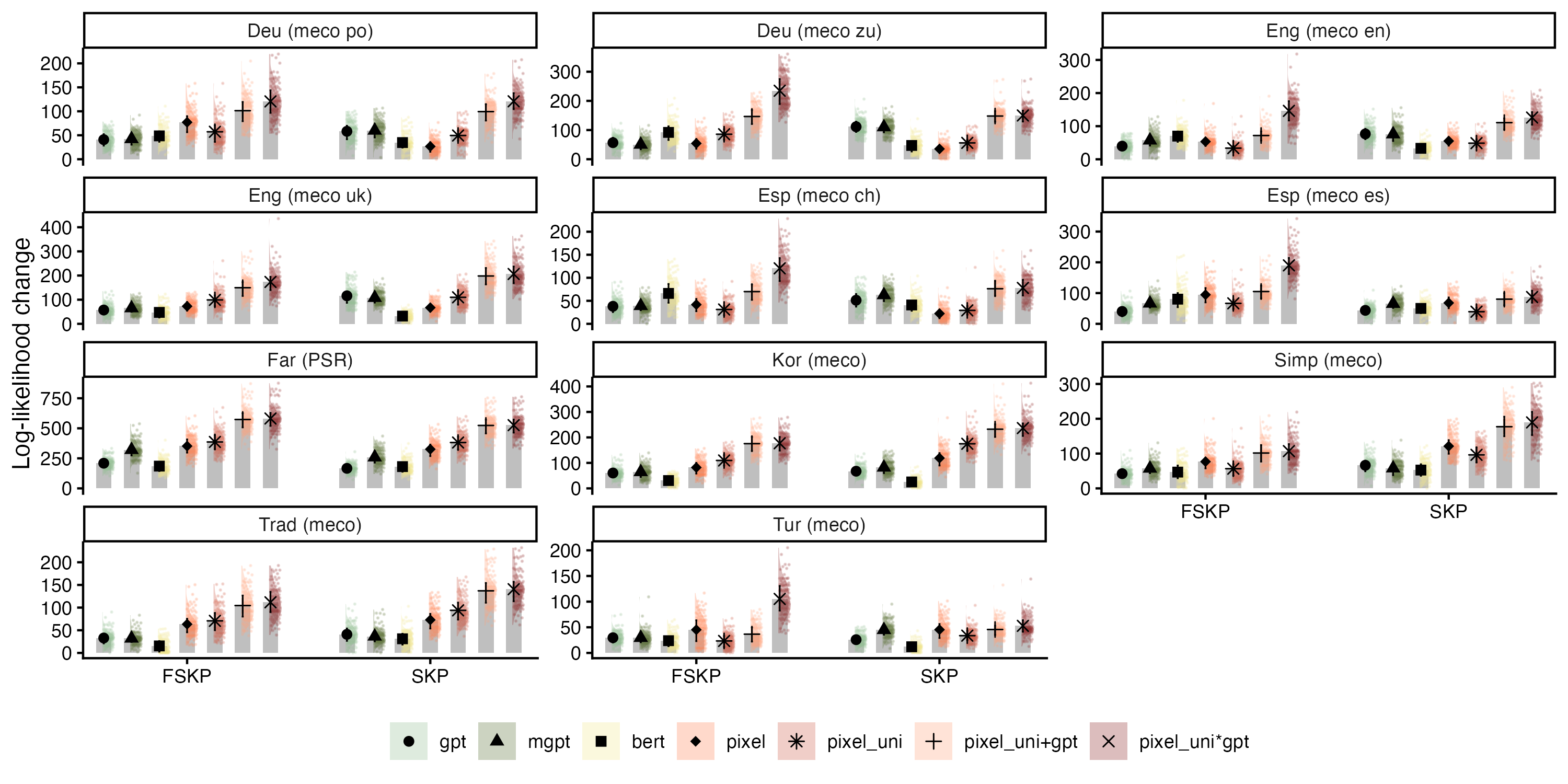}
    \caption{Main results by dataset (Skipping)}
    \label{fig:detailresultskiping}
\end{figure*}
\FloatBarrier
\section{Cross-validation}

\begin{figure*}[t]
    \centering
    \includegraphics[width=1.0\linewidth]{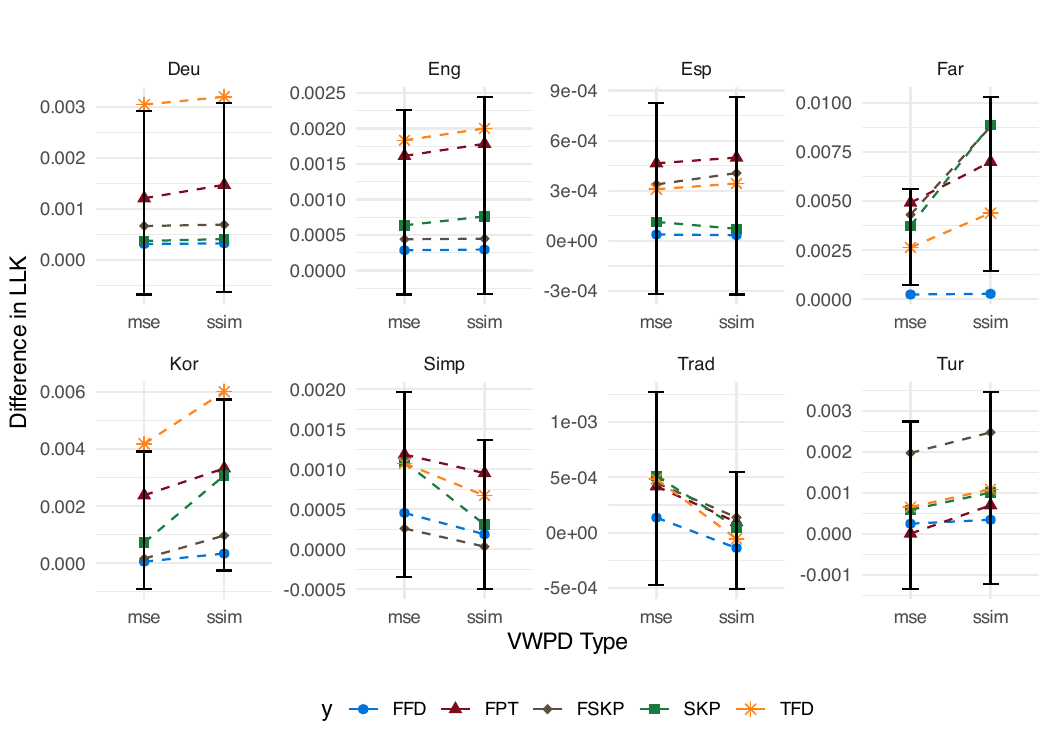}
    \caption{Comparison of MSE and SSIM measures in 10-fold cross-validation settings.}
    \label{fig:cvcompare}
\end{figure*}
\begin{figure*}[t]
    \centering
    \includegraphics[width=\linewidth]{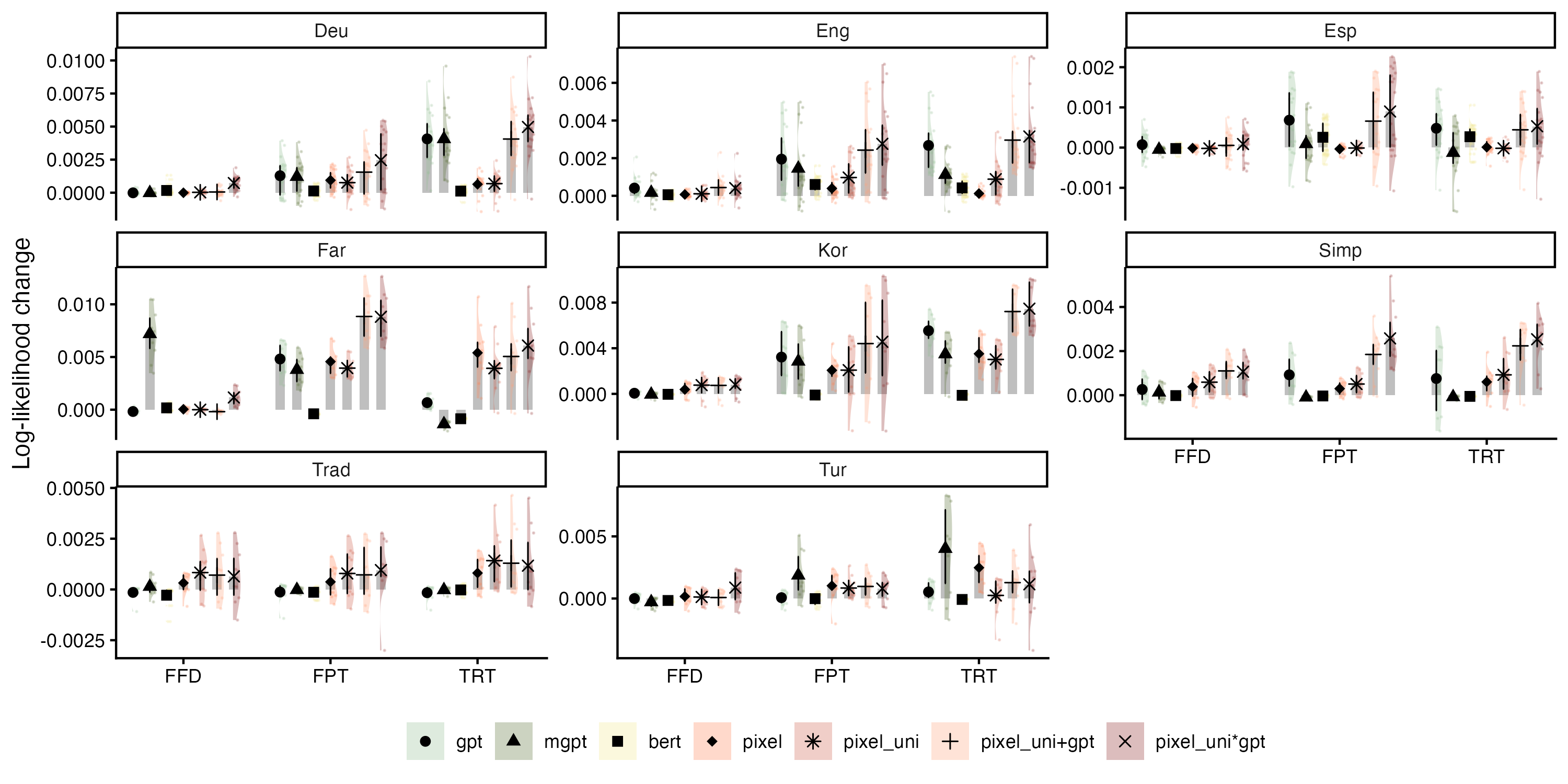}
    \caption{Main results by dataset (Reading Time) using 10-fold cross-validation based on visual and token surprisal. (Log-likelihood change is case-wise averaged)}
    \label{fig:cvresultreadingtime_surprisal}
\end{figure*}
\begin{figure*}[t]
    \centering
    \includegraphics[width=\linewidth]{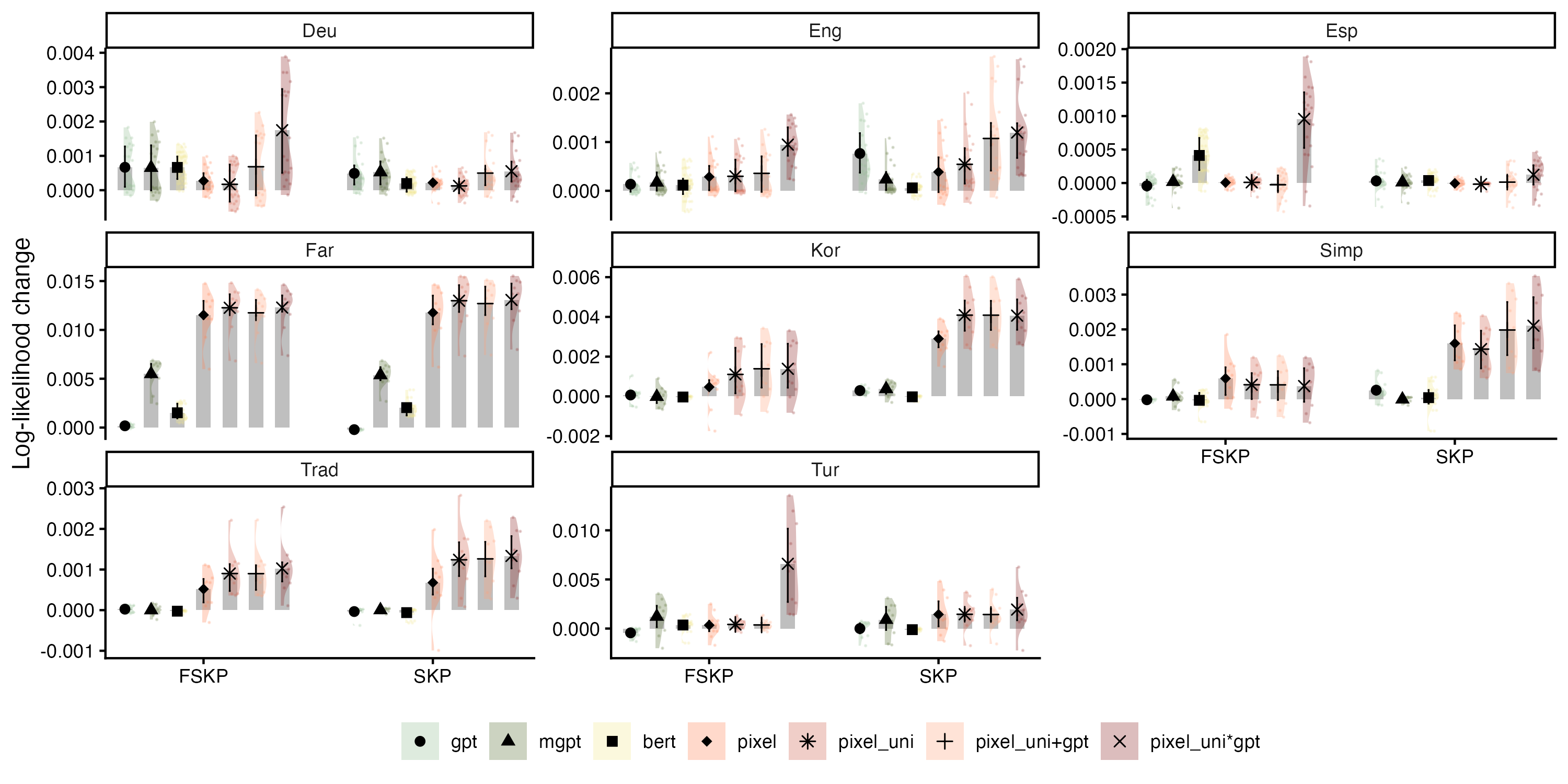}
    \caption{Main results by dataset (Skipping) using 10-fold cross-validation based on visual and token surprisal. (Log-likelihood change is case-wise averaged)}
    \label{fig:cvresultskiping_surprisal}
\end{figure*}
\begin{figure*}[t]
    \centering
    \includegraphics[width=\linewidth]{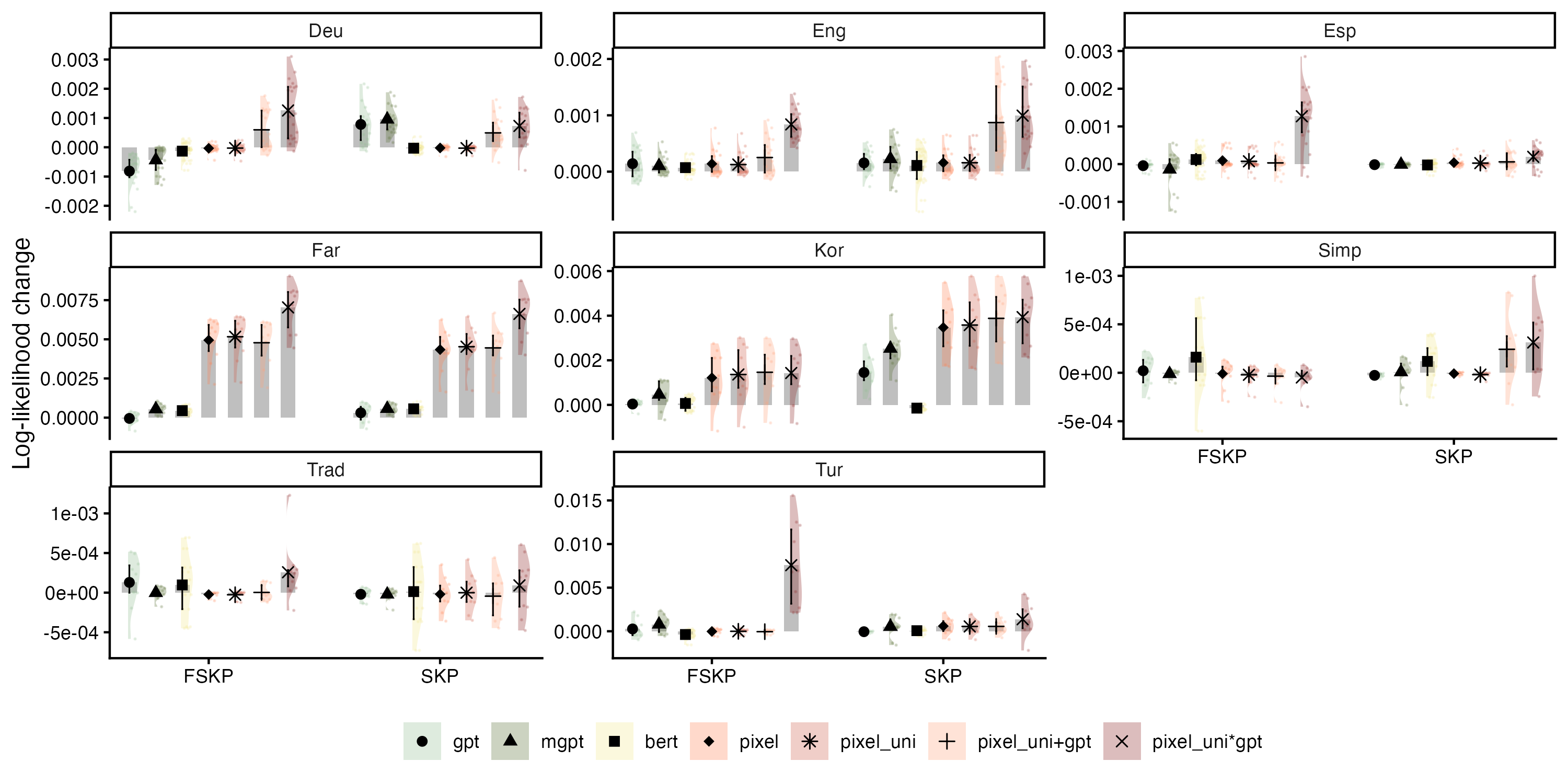}
    \caption{Main results by dataset (Skipping) using 10-fold cross-validation, based on visual and token entropy. (Log-likelihood change is case-wise averaged)}
    \label{fig:cvresultskiping_entropy}
\end{figure*}

\label{app:cv}
In Figure \ref{fig:cvcompare}, we reran the inter-measure comparison results in Figure \ref{fig:vwpcomparison} through 10-fold cross-validation.
In Figure \ref{fig:cvresultreadingtime_surprisal}, Figure \ref{fig:cvresultskiping_surprisal} and Figure \ref{fig:cvresultskiping_entropy}, we reproduce the results in Figure \ref{fig:plot_surprisal_rt}, Figures \ref{fig:plot_surprisal_skp} and \ref{fig:entropy}, and Table~\ref{tab:compare_ssim_mse_cv}.

The cross-validation results do not contradict the observation that the MECO Chinese data show a stronger advantage of MSE-based VWPD. A revisit of the 10-fold cross-validation result supports the identical tendencies we have captured in Section \ref{sec:results}. In the first place, an integration, especially an interaction, between token and visual prediction improves reading-time and skipping modeling. Besides, visual prediction accounts for skipping better than token prediction in certain scripts, notably Persian (Far), Korean, and Chinese scripts.

\section{Comparing MSE and SSIM across datasets}
\label{app:mse_ssim}
We report the performance of MSE and SSIM in terms of their advantage against the GPT2 model in Table \ref{tab:compare_ssim_mse}.

\FloatBarrier

\newcommand{\sym}[1]{\ifmmode^{#1}\else\(^{#1}\)\fi}
\renewcommand{\sym}[1]{\ifmmode^{#1}\else\(^{#1}\)\fi}
\providecommand{\sym}[1]{\ifmmode^{#1}\else(^{#1})\fi}
\begin{table*}[t]
\centering
\small
\setlength{\tabcolsep}{0.01cm}
\begin{tabular}{ll*{11}{c}}
\toprule
VWPD & Measure & Esp (es) & Esp (ch) & Deu (po) & Deu (zu) & Eng (uk) & Eng (en) & Kor & Far (PSR) & Simp & Trad & Tur \\
\midrule
\multicolumn{13}{l}{\textit{pixel\_uni vs.\ gpt}} \\
\midrule
SSIM & FFD & -23.0\sym{***} & -28.3\sym{***} & -32.7\sym{***} & -48.4\sym{***} & -68.9\sym{***} & -30.6\sym{***} & -5.6\sym{*} & -126.6\sym{***} & -27.5\sym{***} & -24.3\sym{***} & -4.1\sym{*} \\
 & FPT & -37.3\sym{***} & -74.3\sym{***} & -60.1\sym{***} & -84.1\sym{***} & -100.4\sym{***} & -49.9\sym{***} & -44.8\sym{***} & -321.2\sym{***} & -63.8\sym{***} & -24.0\sym{***} & -13.2\sym{***} \\
 & TRT & -20.8\sym{***} & -66.7\sym{***} & -214.0\sym{***} & -235.6\sym{***} & -194.3\sym{***} & -94.6\sym{***} & -68.1\sym{***} & -129.1\sym{***} & -90.6\sym{***} & -30.9\sym{***} & -15.5\sym{***} \\
 & FSKP & +26.4\sym{***} & -6.9\sym{**} & +16.4\sym{***} & +28.9\sym{***} & +41.8\sym{***} & -6.1\sym{*} & +49.2\sym{***} & +177.0\sym{***} & -23.1\sym{***} & -16.1\sym{***} & -6.3\sym{***} \\
 & SKP & -4.3\sym{.} & -22.1\sym{***} & -8.7\sym{***} & -55.4\sym{***} & -6.2 & -28.0\sym{***} & +107.6\sym{***} & +214.9\sym{***} & -45.6\sym{***} & -21.1\sym{***} & +7.8\sym{***} \\
\addlinespace
MSE & FFD & -20.3\sym{***} & -21.0\sym{***} & -25.9\sym{***} & -30.1\sym{***} & -75.6\sym{***} & -34.3\sym{***} & -31.4\sym{***} & -140.0\sym{***} & -13.9\sym{***} & -0.9 & -11.2\sym{***} \\
 & FPT & -33.7\sym{***} & -63.9\sym{***} & -73.6\sym{***} & -94.2\sym{***} & -126.2\sym{***} & -72.4\sym{***} & -73.0\sym{***} & -408.7\sym{***} & -49.9\sym{***} & -1.5 & -12.4\sym{***} \\
 & TRT & -23.8\sym{***} & -70.3\sym{***} & -250.2\sym{***} & -257.0\sym{***} & -268.6\sym{***} & -128.4\sym{***} & -123.7\sym{***} & -249.6\sym{***} & -52.9\sym{***} & +39.2\sym{***} & -22.6\sym{***} \\
 & FSKP & +22.6\sym{***} & -4.6\sym{.} & +10.3\sym{**} & +7.3\sym{*} & +16.1\sym{***} & -0.3 & -14.2\sym{***} & -109.4\sym{***} & +14.2\sym{***} & +37.9\sym{***} & -14.2\sym{***} \\
 & SKP & -9.1\sym{***} & -21.7\sym{***} & -14.5\sym{***} & -63.9\sym{***} & -68.1\sym{***} & -35.8\sym{***} & -40.5\sym{***} & -70.7\sym{***} & +30.0\sym{***} & +53.0\sym{***} & -8.1\sym{***} \\
\midrule
\multicolumn{13}{l}{\textit{pixel\_uni $\times$ gpt vs.\ gpt}} \\
\midrule
SSIM & FFD & +26.6\sym{***} & +13.7\sym{***} & +31.7\sym{***} & +37.0\sym{***} & +18.1\sym{***} & +12.1\sym{***} & +21.7\sym{***} & +103.8\sym{***} & +11.1\sym{***} & +4.0\sym{***} & -6.0\sym{*} \\
 & FPT & +20.2\sym{***} & +28.7\sym{***} & +55.9\sym{***} & +78.9\sym{***} & +76.5\sym{***} & +53.0\sym{***} & +36.2\sym{***} & +181.7\sym{***} & +37.0\sym{***} & +34.7\sym{***} & -17.1\sym{***} \\
 & TRT & +43.1\sym{***} & +26.6\sym{***} & +74.4\sym{***} & +65.9\sym{***} & +100.3\sym{***} & +54.2\sym{***} & +59.8\sym{***} & +228.4\sym{***} & +19.3\sym{***} & +7.3\sym{***} & -4.4\sym{*} \\
 & FSKP & +134.7\sym{***} & +49.0\sym{***} & +76.1\sym{***} & +143.9\sym{***} & +65.1\sym{***} & +56.2\sym{***} & +114.5\sym{***} & +328.7\sym{***} & +16.2\sym{***} & +23.7\sym{***} & +16.1\sym{***} \\
 & SKP & +42.5\sym{***} & +16.9\sym{***} & +43.0\sym{***} & +19.1\sym{***} & +85.4\sym{***} & +42.0\sym{***} & +167.4\sym{***} & +337.6\sym{***} & +19.2\sym{***} & +23.7\sym{***} & +15.7\sym{***} \\
\addlinespace
MSE & FFD & +30.6\sym{***} & +21.8\sym{***} & +34.8\sym{***} & +45.0\sym{***} & +16.1\sym{***} & +8.7\sym{***} & -0.7 & +56.2\sym{***} & +33.4\sym{***} & +28.9\sym{***} & -11.1\sym{***} \\
 & FPT & +23.1\sym{***} & +37.1\sym{***} & +57.2\sym{***} & +84.9\sym{***} & +64.0\sym{***} & +40.0\sym{***} & +18.4\sym{***} & +57.7\sym{***} & +66.3\sym{***} & +59.9\sym{***} & -14.8\sym{***} \\
 & TRT & +41.2\sym{***} & +21.4\sym{***} & +64.6\sym{***} & +73.6\sym{***} & +51.8\sym{***} & +31.1\sym{***} & +23.3\sym{***} & +79.2\sym{***} & +78.4\sym{***} & +82.7\sym{***} & -10.8\sym{***} \\
 & FSKP & +132.7\sym{***} & +54.0\sym{***} & +66.1\sym{***} & +112.0\sym{***} & +36.8\sym{***} & +56.1\sym{***} & +44.2\sym{***} & +56.2\sym{***} & +61.0\sym{***} & +72.2\sym{***} & +5.6\sym{*} \\
 & SKP & +37.1\sym{***} & +19.5\sym{***} & +41.9\sym{***} & +27.5\sym{***} & +35.4\sym{***} & +46.6\sym{***} & +31.1\sym{***} & +63.0\sym{***} & +119.1\sym{***} & +97.4\sym{***} & +2.9 \\
\bottomrule
\end{tabular}
\caption{Mean paired log-likelihood difference relative to GPT. Positive values favour the reported model. The estimation is based on boostrapping. The significance is based on paired t-test.\sym{*} $p<.05$, \sym{**} $p<.01$, \sym{***} $p<.001$, \sym{.} $p<.1$. }
\label{tab:compare_ssim_mse}
\end{table*}

\begin{table*}[t]
\centering
\small
\setlength{\tabcolsep}{0.01cm}

\begin{tabular}{ll*{6}{c}}
\toprule
VWPD & Measure & Esp (es) & Esp (ch) & Deu (po) & Deu (zu) & Eng (uk) & Eng (en) \\
\midrule
\multicolumn{8}{l}{\textit{pixel\_uni vs.\ gpt}} \\
\midrule
SSIM & FFD & -0.000033 & -0.000153 & +0.000029 & +0.000045\sym{.} & -0.000421\sym{.} & -0.000188\sym{.} \\
 & FPT & -0.000374 & -0.001012\sym{**} & -0.000885\sym{.} & -0.000158 & -0.001189\sym{.} & -0.000768\sym{**} \\
 & TRT & -0.000125 & -0.000871\sym{***} & -0.004425\sym{***} & -0.002344\sym{***} & -0.002263\sym{***} & -0.001323\sym{***} \\
 & FSKP & +0.000098 & +0.000007 & -0.000392\sym{***} & -0.000599\sym{***} & +0.000323\sym{*} & +0.000003 \\
 & SKP & -0.000009 & -0.000087 & -0.000167\sym{.} & -0.000562\sym{***} & -0.000131 & -0.000313\sym{**} \\
\addlinespace
MSE & FFD & -0.000036 & -0.000146\sym{.} & +0.000089 & +0.000148 & -0.000500\sym{*} & -0.000215\sym{*} \\
 & FPT & -0.000376\sym{.} & -0.001025\sym{**} & -0.001090\sym{*} & -0.001178\sym{*} & -0.001984\sym{**} & -0.001001\sym{**} \\
 & TRT & -0.000134 & -0.000860\sym{***} & -0.005057\sym{***} & -0.002882\sym{***} & -0.003238\sym{***} & -0.001759\sym{***} \\
 & FSKP & +0.000123 & +0.000002 & -0.000342\sym{**} & -0.001444\sym{***} & +0.000232\sym{.} & +0.000027 \\
 & SKP & -0.000004 & -0.000005 & -0.000246\sym{*} & -0.000756\sym{**} & -0.000857\sym{***} & -0.000382\sym{**} \\
\midrule
\multicolumn{8}{l}{\textit{pixel\_uni $\times$ gpt vs.\ gpt}} \\
\midrule
SSIM & FFD & -0.000008 & +0.000003 & +0.000429\sym{*} & +0.000408\sym{.} & -0.000064 & -0.000207\sym{.} \\
 & FPT & +0.000040 & +0.000230\sym{*} & +0.000511 & +0.001832\sym{***} & +0.001246\sym{***} & +0.000365\sym{***} \\
 & TRT & -0.000155 & +0.000232\sym{**} & +0.000863\sym{*} & +0.000743\sym{*} & +0.000505 & +0.000352\sym{*} \\
 & FSKP & +0.001094\sym{***} & +0.000184 & +0.000259\sym{*} & +0.001035\sym{***} & +0.000131 & +0.000318\sym{**} \\
 & SKP & +0.000186\sym{**} & -0.000164\sym{*} & -0.000454\sym{*} & -0.000338\sym{***} & +0.000578\sym{**} & +0.000175\sym{*} \\
\addlinespace
MSE & FFD & -0.000006 & +0.000027 & +0.000397\sym{**} & +0.000283 & -0.000072 & -0.000219\sym{.} \\
 & FPT & -0.000005 & +0.000108 & +0.000677 & +0.001291\sym{**} & +0.000839\sym{***} & +0.000222\sym{**} \\
 & TRT & -0.000179 & +0.000171\sym{.} & +0.000861\sym{*} & +0.000642\sym{*} & +0.000021 & +0.000042 \\
 & FSKP & +0.001180\sym{***} & +0.000225 & +0.000162 & +0.000210 & +0.000030 & +0.000331\sym{**} \\
 & SKP & +0.000169\sym{**} & -0.000063 & -0.000309\sym{.} & -0.000299\sym{**} & +0.000060 & +0.000258\sym{**} \\
\bottomrule
\end{tabular}
\par\medskip

\begin{tabular}{ll*{5}{c}}
\toprule
VWPD & Measure & Kor & Far (PSR) & Simp & Trad & Tur \\
\midrule
\multicolumn{7}{l}{\textit{pixel\_uni vs.\ gpt}} \\
\midrule
SSIM & FFD & +0.000710\sym{.} & +0.000173 & -0.000287 & +0.000074 & +0.000127 \\
 & FPT & -0.001143\sym{.} & -0.000866 & -0.000944\sym{**} & +0.000049 & +0.000777\sym{.} \\
 & TRT & -0.002521\sym{**} & +0.003256\sym{***} & -0.000795 & +0.000122 & -0.000284 \\
 & FSKP & +0.001036\sym{.} & +0.012088\sym{***} & -0.000023 & -0.000052 & +0.000854\sym{*} \\
 & SKP & +0.003796\sym{***} & +0.013201\sym{***} & -0.000286\sym{*} & -0.000024 & +0.001453\sym{*} \\
\addlinespace
MSE & FFD & -0.000061 & +0.000217 & +0.000329 & +0.000978\sym{*} & +0.000007 \\
 & FPT & -0.003004\sym{**} & -0.003013\sym{***} & -0.000419 & +0.000917\sym{.} & -0.000087 \\
 & TRT & -0.005010\sym{***} & +0.000656\sym{*} & +0.000167 & +0.001579\sym{**} & -0.000645\sym{*} \\
 & FSKP & +0.000079 & +0.002944\sym{***} & +0.000431\sym{*} & +0.000877\sym{***} & +0.000464\sym{*} \\
 & SKP & +0.000351 & +0.003176\sym{***} & +0.001179\sym{***} & +0.001274\sym{**} & +0.000643\sym{.} \\
\midrule
\multicolumn{7}{l}{\textit{pixel\_uni $\times$ gpt vs.\ gpt}} \\
\midrule
SSIM & FFD & +0.000555 & +0.001022\sym{***} & -0.000181\sym{.} & -0.000408 & +0.000556 \\
 & FPT & +0.001384\sym{.} & +0.003938\sym{***} & +0.000578 & +0.000228 & +0.000777\sym{*} \\
 & TRT & +0.001847\sym{**} & +0.004196\sym{**} & +0.000071 & +0.000149 & +0.000076 \\
 & FSKP & +0.001260\sym{*} & +0.009560\sym{***} & -0.000053 & +0.000180 & +0.003422\sym{**} \\
 & SKP & +0.003754\sym{***} & +0.010907\sym{***} & -0.000003 & +0.000073 & +0.001872\sym{*} \\
\addlinespace
MSE & FFD & -0.000212\sym{.} & +0.000435\sym{*} & +0.000687\sym{*} & +0.000520 & +0.000395 \\
 & FPT & +0.000341 & +0.000923\sym{**} & +0.001424\sym{**} & +0.000952 & -0.000039 \\
 & TRT & +0.000694\sym{**} & +0.000879\sym{*} & +0.001578\sym{**} & +0.001596\sym{.} & -0.000039 \\
 & FSKP & +0.000048 & +0.001996\sym{***} & +0.000391 & +0.000988\sym{***} & +0.003323\sym{*} \\
 & SKP & +0.000702\sym{***} & +0.002047\sym{***} & +0.001822\sym{***} & +0.001355\sym{***} & +0.001367\sym{*} \\
\bottomrule
\end{tabular}
\caption{Mean paired log-likelihood difference relative to GPT. Positive values favour the reported model. \sym{*} $p<.05$, \sym{**} $p<.01$, \sym{***} $p<.001$, \sym{.} $p<.1$. The estimations are based on cross-validation.  The significance is based on paired t-test.}
\label{tab:compare_ssim_mse_cv}
\end{table*}

\end{document}